\documentclass[nojss]{jss}

\usepackage[utf8]{inputenc}
\usepackage{color}

\usepackage{amssymb}
\usepackage{amsmath}
\usepackage{txfonts}
\usepackage{mathdots}
\usepackage{float}
\usepackage{algorithm}
\usepackage{algorithmic}
\usepackage{booktabs}
\usepackage{multirow}
\usepackage{tablefootnote}
\usepackage{longtable}
\usepackage{tabularx}
\usepackage{parnotes}
\usepackage{thumbpdf}
\usepackage{caption}
\usepackage{bbm}

\usepackage{letltxmacro}

\newlength{\commentindent}
\makeatletter
\renewcommand{\algorithmiccomment}[1]{\unskip\hfill\makebox[\commentindent][l]{//~#1}\par}
\LetLtxMacro{\oldalgorithmic}{\algorithmic}
\renewcommand{\algorithmic}[1][0]{%
  \oldalgorithmic[#1]%
  \renewcommand{\ALC@com}[1]{%
    \ifnum\pdfstrcmp{##1}{default}=0\else\algorithmiccomment{##1}\fi}%
}
\makeatother

\DeclareMathOperator*{\argmax}{arg\,max}

\usepackage{natbib}
\usepackage[british]{babel} % for correct word hyphenation
\author{Robin van Emden\\JADS \And
  Maurits Kaptein\\Tilburg University}

\title{\pkg{contextual}: Evaluating Contextual Multi-Armed Bandit Problems in R}

\Plainauthor{Robin van Emden, Maurits Kaptein} %% comma-separated
\Plaintitle{contextual: Evaluating Contextual Multi-Armed Bandit Problems in R} %% without formatting
\Shorttitle{\pkg{contextual}} %% a short title (if necessary)

\Abstract{

Over the past decade, contextual bandit algorithms have been gaining in popularity due to their effectiveness and flexibility in solving sequential decision problems---from 
online advertising and finance to clinical trial design and personalized medicine. At the same time, there are, as of yet, surprisingly few options that enable researchers 
and practitioners to simulate and compare the wealth of new and existing bandit algorithms in a standardized way. To help close this gap between analytical research and 
empirical evaluation the current paper introduces the object-oriented \proglang{R} package \pkg{contextual}: a user-friendly and, through its object-oriented design, easily 
extensible framework that facilitates parallelized comparison of contextual and context-free bandit policies through both simulation and offline analysis.
}

\Plainkeywords{contextual multi-armed bandits, simulation, sequential experimentation, R}

\Address{
  Robin van Emden\\
  Jheronimus Academy of Data Science\\
  Den Bosch, the Netherlands\\
  E-mail: \email{robinvanemden@gmail.com} \\
  URL: \url{pavlov.tech}\\
  \linebreak
  Maurits C. Kaptein\\
  Tilburg University\\
  Statistics and Research Methods\\
  Tilburg, the Netherlands\\
  E-mail: \email{m.c.kaptein@uvt.nl}\\
  URL: \url{www.mauritskaptein.com}\\
}

\begin{document}
%%\SweaveOpts{concordance=TRUE}
\sloppy

%% A vignette for the \cite{contextual} paper. #########################################

%% include your article here, just as usual
%% Note that you should use the \pkg{}, \proglang{} and \code{} commands.

\section{Introduction} \label{intro}

There are many real-world situations in which we have to decide between multiple options, yet are only able to learn the best course of action by testing each option 
sequentially. In such situations, the underlying concept remains the same for each and every renewed decision: Do you stick to what you know and receive an expected result 
("exploit") or choose an option you do not know all that much about and potentially learn something new ("explore")? As we all encounter such dilemma's on a daily basis 
\citep{Wilson2014}, it is easy to come up with examples - for instance:

\begin{itemize}
\item When going out to dinner, do you explore new restaurants, or choose a favorite?
\item As a website editor, do you place popular or new articles at the top of your frontpage?
\item As a doctor, do you prescribe tried and tested medication, or do you also provide promising experimental drugs?
\item When visiting a casino, do you stay with the slot machine that just paid out, or do you try some of the other slot machines?
\end{itemize}

Although people seem to navigate such explore-exploit problems with relative ease, this type of decision problem has proven surprisingly difficult to solve 
analytically\footnote{As Dr. Peter Whittle famously stated "[the problem] was formulated during the [second world] war, and efforts to solve it so sapped the energies and 
minds of Allied analysts that the suggestion was made that the problem be dropped over Germany, as the ultimate instrument of intellectual sabotage." \citep{Whittle1979}} 
and has been studied extensively since the 1930s \citep{Robbins1952,Bubeck2012} under the umbrella of the "multi-armed bandit" (MAB) problem. The origin of the name is 
related to the casino example above: a one armed bandit is an old name for a slot machine in a casino, as they used to have one arm and tended to steal your money. A 
multi-armed bandit can then be understood as a set of one-armed bandit slot machines in a casino---in that respect, "many one-armed bandits problem" might have been a better 
fit \citep{2018a}. Just like in the casino example, the crux of a multi-armed bandit problem is that you only receive a reward for the arm you pull---you remain in the dark 
about what rewards the other arms might have offered. Consequently, you need some strategy or "policy" that helps you balance the exploration and exploitation of arms to 
optimize your rewards over repeated pulls. One option would, for instance, be to pull every available arm once and from then on exploit the arm that offered you the highest 
reward. This reward might, however, be nothing more than a lucky fluke. On the other hand, if you decide to keep exploring other arms, you may lose out on the winnings you 
might have received from the arm that had been doing so well.

This active exploration of information gathered one step at a time makes multi-armed bandit problems a subset of reinforcement learning type problems ---together with 
supervised and unsupervised learning one of three main classes of machine learning. Where supervised algorithms learn mappings from input values to fully specified class 
labels and unsupervised learning looks for patterns in data without any such labels, reinforcement learning policies live somewhere in between: they are able to make use of 
labels or "rewards" to minimize their loss---but in contast to their supervised siblings, this information has to be actively acquired \citep{Jordan2015}. Bandit policies 
then constitute that subset of reinforcement learning algorithms that either do not take further contextual information into account or, otherwise, assume that their choices 
do not affect this context \citep{steenwinckel2018self,Sutton1998e}.

Where the latter brings us to a MAB generalization generally known as the \textit{contextual} multi-armed bandit (CMAB) problem. CMAB problems extend on basic "context-free" 
MABs by adding one crucial element: contextual information \citep{Langford2008}. Contextual multi-armed bandits are known by many different names in about as many different 
fields of research \citep{Tewari2017}---for example as "bandit problems with side observations" \citep{Wang2005a}, "bandit problems with side information" \citep{Lu2010}, 
"associative reinforcement learning" \citep{Kaelbling1996}, "reinforcement learning with immediate reward" \citep{Abe2003}, "associative bandit problems" \citep{Strehl2006}, 
or "bandit problems with covariates" \citep{Sarkar1991}. However, the term "contextual multi-armed bandit," as conceived by \cite{Langford2008}, is the most used---so that 
is the term we will use in the current paper.

However named, in contextual bandit problems, CMAB policies differentiate themselves, by definition, from their MAB cousins in that they are able to make use of features 
that reflect the current state of the world---features that can then be mapped onto available arms or actions\footnote{That is, before making a choice, the learner receives 
information on the state of the world or "context" in the form of a d-dimensional feature vector. After making a choice the learner is then able to combine this contextual 
information with the reward received to make a more informed decision in the next round.}. This access to side information makes CMAB algorithms yet more relevant to many 
real-life decision problems than their MAB progenitors \citep{Langford2008}. To follow up on our previous examples: do you choose the same  type of restaurants in your 
hometown and when on vacation? Do you prescribe the same treatment to male and female patients? Do you place the same news story on the frontpage of your website for both 
young and old visitors? Probably not---it makes sense to make use of any additional contextual information that can help you make a better decision. So it may be no surprise 
that CMAB algorithms have found applications in many different areas: from recommendation engines \citep{Lai1985} to advertising \citep{Tang2013} and (personalized) medicine 
\citep{Katehakis1986,Tewari2017}, healthcare \citep{Rabbi2015}, and portfolio choice \citep{Shen2015}---inspiring a multitude of new bandit algorithms or policies.

However, although CMAB algorithms have found more and more applications, comparisons on both synthetic, and, importantly, real-life, large-scale offline datasets 
\citep{Li2011} have relatively lagged behind\footnote{Here, a \textbf{synthetic} data generator (or \code{Bandit}, in \pkg{contextual} parlance) compares policies against 
some simulated environment, usually seeking to model or emulate some online bandit scenario---whereas an \textbf{offline} \code{Bandit} compares policies against a previous 
collected data set---generally logged with a completely different policy than the one(s) under evaluation \citep{Li2012}.}. To this end, the current paper introduces the 
\proglang{R} package \pkg{contextual}, to facilitate the development, evaluation, and comparison the of (contextual) multi-armed bandit policies by offering an easily 
extensible, class-based, modular architecture.

In that respect, \pkg{contextual} differentiates itself from several other types of bandit oriented software applications and services, such as:

\begin{enumerate}
          \item[1)]Online A/B and basic, out-of-the-box MAB test services such as \pkg{Google Analytics} \citep{BibEntry2018Aug2}, \pkg{Optimizely} \citep{BibEntry2018Aug3}, 
	  \pkg{Mix Panel} \citep{BibEntry2018Aug1}, \pkg{AB Tasty} \citep{BibEntry2018AugA}, \pkg{Adobe Target} \citep{BibEntry2018AugB}, and more.
          \item[2)]More advanced online CMAB test services and software, such as the flexible online evaluation platform \pkg{StreamingBandit} 
	  \citep{kruijswijk2018streamingbandit} and Microsoft's \pkg{Custom Decision Service} \citep{Agarwal2016}.
          \item[3)]Predominantly context-free simulation oriented projects such as \pkg{Yelp MOE} \citep{2018}, which runs sequential A/B tests using Bayesian optimization, 
	  and the mainly MAB focused \proglang{Python} packages \pkg{Striatum} \citep{striatum} and \pkg{SMPyBandits} \citep{SMPyBandits}.
          \item[4)]Software that facilitates the evaluation of bandit policies on offline data, such as \pkg{Vowpal Wabbit} \citep{Langford2007}, \pkg{Jubatus} 
	  \citep{Hido2013}, and \pkg{TensorFlow} \citep{Abadi2016}.
\end{enumerate}

Though each of these applications and services may share certain features with \pkg{contextual}, overall, \pkg{contextual} clearly distinguishes itself in several respects. 
 First, it focusses on the evaluation of bandit policies on simulated and offline datasets, which discriminates it from the online evaluation oriented packages listed under 
items 1 and 2. Second, though \pkg{contextual} is perfectly capable of simulating and comparing context-free MAB policies, its emphasis lies on the simulation of contextual 
policies, distinguishing it from the projects listed under item 3. Finally, though \pkg{contextual} is closely related to the projects listed under item 4, it also, again, 
differentiates itself in several key respects:

\begin{enumerate}
          \item[a)]\pkg{contextual} offers a diverse, open and extensible library of common MAB and CMAB policies.
          \item[b)]\pkg{contextual} is developed in \proglang{R}, opening the door to a lively exchange of code, data, and knowledge between scientists and practitioners 
	  trained in \proglang{R}.
          \item[c)]\pkg{contextual} focusses on ease of conversion of existing and new algorithms into clean, readable and shareable source code.
          \item[d)]In building on \proglang{R}'s \pkg{doParallel} package, \pkg{contextual}'s simulations are parallelized by default---and can easily be run on different 
	  parallel architectures, from cluster (such as on Microsoft Azure, Amazon ec2 or Hadoop backends) to GPU based.
\end{enumerate}

All in all, though there are some alternatives, there was, as of yet, no extensible and widely applicable \proglang{R} package to analyze and compare, respectively, basic 
multi-armed, continuum \citep{Agrawal1995} and contextual multi-armed bandit algorithms on both simulated and offline data---outside of single-use scripts or basic or 
isolated code repositories \citep{Gandrud2016}.

In making our latest CMAB \proglang{R} package \pkg{contextual} openly available at https://github.com/Nth-iteration-labs/contextual, we hope to remedy this situation, 
focussing on two goals:

\begin{enumerate}
          \item[1)]Easing the implementation, evaluation, and dissemination of (C)MAB policies.
          \item[2)]Introducing a wider audience to (C)MAB  policies' advanced sequential decision strategies.
\end{enumerate}

The current paper pays heed to both these goals, in introducing the \pkg{contextual} package through basic CMAB simulations, running simple policies on bandits with just a 
few arms---leading up to a partial replication of \cite{Li2010} that puts all introduced elements together in a demonstration of how \pkg{contextual} is able to efficiently 
compare seven policies in parallel on 45,811,883 separate events with a continually shifting pool of active arms for six levels of sparsity, therein uncovering some 
interesting new findings, all within 24 hours (see Section \ref{repl}, Figure \ref{fig:section_8_plot}).

Specifically, Section \ref{formalizationandimplementation} starts out by presenting a formal definition of the contextual multi-armed bandit problem, shows how this 
formalization can be transformed into a clear and concise object-oriented architecture, and then describes how to set up a minimal simulation. Section \ref{basicusage} gives 
an overview of \pkg{contextual}'s predefined \code{Bandit} and \code{Policy} subclasses and demonstrates how to run a very basic \code{Simulation}. Section 
\ref{classstructure} delves a little deeper into the implementation of each of \pkg{contextual}'s core superclasses. Section \ref{extending} shows how to extend 
\pkg{contextual}'s superclasses to create your own custom \code{Bandit} and \code{Policy} subclasses. Section \ref{subclpb} demonstrates how to further subclass existing 
\code{Bandit} and \code{Policy} implementations. Section \ref{offl} focusses on how a \code{Bandit} subclass can make use of offline datasets. Section \ref{repl} brings all 
of the previous sections together in a partial replication of a frequently cited contextual bandit paper. In Section \ref{future} we conclude with some comments on the 
current state of the package and potential future enhancements.

This organization offers the reader several ways to peruse the current paper. Readers with a passing knowledge of \proglang{R} who are seeking to run simulations based on 
\pkg{contextual}'s default bandits and policies should be able to get up and running by reading the current introduction plus Section \ref{basicusage}. For a more formal 
introduction, include Section \ref{formalizationandimplementation}. For readers who know their way around \proglang{R} and who are seeking to extend \pkg{contextual} to run 
custom bandits and policies, it is probably best to read the whole paper---with a focus on Sections \ref{classstructure}, \ref{extending} and \ref{subclpb}. Finally, add 
Sections \ref{repl} and possibly \ref{future} for those readers interested in the implementation of custom offline bandits.

\section{Formalization and implementation} \label{formalizationandimplementation}

In the current section, we first introduce a more formal definition of the contextual multi-armed bandit problem. Next, we present our concise implementation and demonstrate 
how to put together a minimal MAB simulation.

\subsection{Formalization} \label{formalization}

\subsubsection{Bandit} \label{formalization}

Bandit $B$ can be defined as a set of arms \(k \in \left\{ 1, \dots, K \right\}\), where each arm is described by some reward function mapping $d$ dimensional context vector 
$x_{t,k}$ to some reward $r_{t,k}$ \citep{Auer2002,Langford2008} for every time step $t$ until horizon $T$.

\subsubsection{Policy} \label{formalization}

Policy $\piup$ seeks to maximize its cumulative reward $\sum_{t=1}^T r_t$ (or minimize its cumulative regret---see equations \ref{eq:1}, \ref{eq:2}) by sequentually 
selecting one of bandit $B$'s currently available arms \citep{Bubeck2012}, here defined as taking action $a_t$ in $\mathcal{A}_t \subseteq K$ for \emph{t}= \{1, \ldots, 
T\}.\footnote{In other words: Every $t$ a policy makes a choice $a_t$ from a set of $\mathcal{A}_t$ actions, which is often equal to---but can be a subset of---the set of 
$K$ arms of bandit $B$.}

\subsubsection{Arm-selection process} \label{formalization}

At each time step $t$ policy $\piup$ first observes the current state of the world as related to $B$, represented by $d$-dimensional context feature vectors 
\(x_{t,a}\)\footnote{Some formalizations describe this vector as $x_t$ (see, for example, \cite{Slivkins2014} or \cite{May2012}). This formalization fits bandit scenario's 
where contextual information at each $t$ is the same for each arm. For instance, a user feature vector $x_t$ comprising the $d$ features of a visitor to a news site, with 
articles for arms. Yet when arm-specific features are taken into account as well, formalizations generally describe the feature vector at each $t$ as $x_{t,a}$ (e.g. 
\cite{Chu2009}, \cite{Li2010})---thereby summarizing, for example, information on both a user $u_t$ and article $k$. To facilitate both approaches, \pkg{contextual} expects 
its bandits to generate a $d \times k$ dimensional context feature matrix $X_t$ at each $t$.} for \(a_{t} \in \mathcal{A_t}\). Making use of some arm-selection strategy, 
policy $\piup$ then selects one of the available actions in $\mathcal{A}_t$. As a result of selecting action $a_t$, policy $\piup$ then receives reward \(r_{a_{t},t}\). With 
observation \( (x_{t,a_t},a_{t},r_{t,a_t}) \), the policy can now update its arm-selection strategy. This cycle is then repeated \textit{T} times.

Additionally, policies generally also summarize historical interactions \( D_{t'} = (x_{t'},a_{t'},r_{t'}) \) over \emph{t}= \{1, \ldots, t'\} by use of a limited set of 
parameters $\theta_{t}$ \citep{kruijswijk2018streamingbandit}. This ensures a policy's action-selection process remains computationally tractable in keeping the 
dimensionality of $\theta_{t'} << D_{t'}$.

\clearpage

Schematically, for each round \emph{t}= \{1, \ldots, T\}:

\begin{enumerate}
         \item[1)] Policy $\piup$ observes current context feature vectors $x_{t,a}$ for $\forall a \in \mathcal{A}_t$ in bandit $B$
         \item[2)] Based on all $x_{t,a}$ and $\theta_{t-1}$, policy $\piup$ now selects an action \(a_{t} \in \mathcal{A}_t\)
         \item[3)] Policy $\piup$ receives a reward \(r_{t,a_t,x_t}\) from bandit $B$
         \item[4)] Policy $\piup$ updates arm-selection strategy parameters $\theta_{t}$ with \( (x_{t,a_t},a_t,r_{t,a_t}) \)
\end{enumerate}

\subsubsection{Goal and performance measure} \label{formalization}

The goal of the policy $\piup$ is to optimize its \textit{cumulative reward} over \emph{t}= \{ 1, \ldots, T \}

\begin{equation} \label{eq:1}
R_{T} = \sum^{T}_{t=1}(r_{t,a_t,x_t})
\end{equation}

The most popular performance measure for bandit policies is \textit{expected cumulative regret} \citep{Kuleshov2014}---defined as the sum of rewards that would have been 
received by choosing optimal action  $a*$ at every \emph{t} subtracted by the sum of rewards awarded to the chosen action $a$ at every \emph{t} over \emph{t}= \{ 1, \ldots, 
T \}:

\begin{equation} \label{eq:2}
\mathbb{E}\left[R_{T} \right] = \mathbb{E}\left[  \max_{a*_t = 1, \dots, \mathcal{A}_t} \sum^{T}_{t=1}(r_{t,a*_t,x_t}) - \sum^{T}_{t=1}(r_{t,a_t,x_t})\right]
\end{equation}

Where expectation $\mathbb{E}\left[ \mathord{\cdot}\right]$ is taken with respect to random draws of both rewards assigned by a bandit and arms as selected by a policy 
\citep{Zheng2016a}.

\subsection{Implementation} \label{implementation}

\begin{figure}[H]
  \centering
    \includegraphics[width=.99\textwidth]{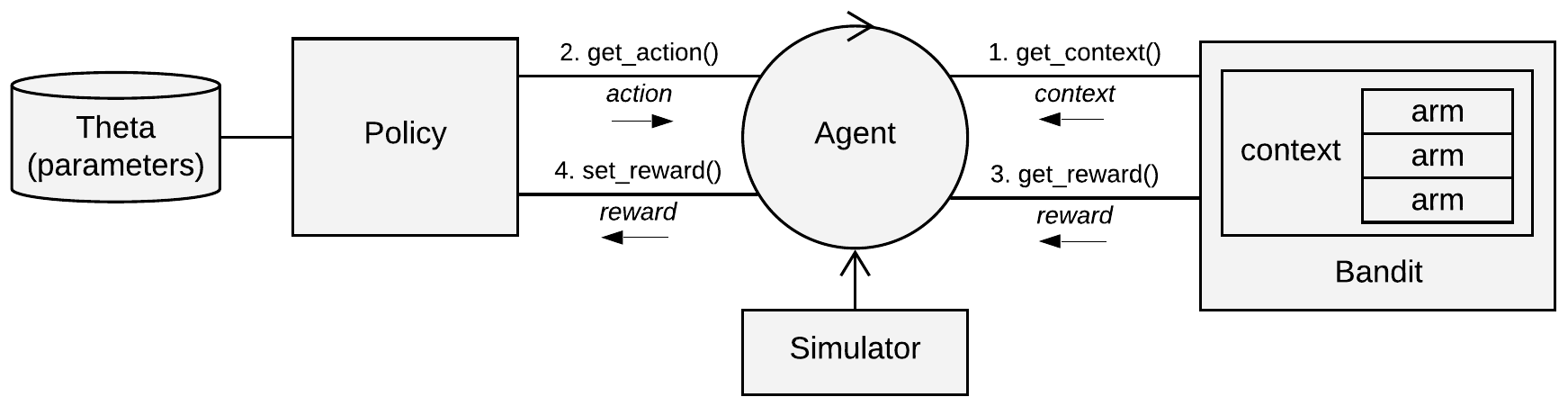}

      \caption{Diagram of \pkg{contextual}'s basic structure. The context feature matrix returned by get\_context() is only taken into account by CMAB policies, and may be 
      ignored by MAB policies.}
      \label{fig:cmab_chart}
\end{figure}

The package's \code{Bandit} and \code{Policy} centered  class structure stays close to the previous formalization, while offering a clean, developer oriented interface. The 
following six classes form the backbone of the package (see also Figure ~\ref{fig:cmab_chart}):

\begin{itemize}
         \item \code{Bandit}: R6 class \code{Bandit} is the parent class of all \pkg{contextual} \code{Bandit} subclasses. It is responsible for the generation of 
	 \code{contexts} and \code{rewards}.

         \item \code{Policy}: R6 class \code{Policy} is the parent class of all \pkg{contextual}'s \code{Policy} implementations. For each \emph{t} = \{1, \ldots, T\} it has 
	 to choose one of a \code{Bandit}'s \code{k} arms, and update its parameters \code{theta} in response to the resulting reward.

         \item \code{Agent}: R6 class \code{Agent} is responsible for the running of one \code{Bandit}/\code{Policy} pair. As such, multiple \code{Agent}s can be run in 
	 parallel with each separate Agent keeping track of \code{t} for its assigned \code{Policy} and \code{Bandit} pair. To be able to fairly evaluate and compare each 
	 agent's performance, and to make sure that simulations are replicable, seeds are set equally and deterministically for each agent over all \code{horizon} times 
	 \code{simulations} time steps of each agent's simulation.

         \item \code{Simulator}: R6 class \code{Simulator} is the entry point of any \pkg{contextual} simulation. It encapsulates one or more \code{Agents} (running in 
	 parallel to each other, by default), creates an \code{Agent} clone (each with its own deterministic seed) for each to be repeated simulation, runs the 
	 \code{Agents}, and saves the log of all \code{Agent} interactions to a \code{History} object.

         \item \code{History}: R6 class \code{History} keeps a log of all \code{Simulator} interactions. It allows several ways to interact with the data, provides summaries 
	 of the data, and can save and load simulation data in several different (\code{data.table}, \code{data.frame} and CSV) formats.

         \item \code{Plot}: R6 class \code{Plot} generates plots from \code{History} logs. It is usually invoked by calling the generic \code{plot(h)} function, where 
	 \code{h} is an \code{History} class instance.
\end{itemize}

\subsection{Putting it together: a first MAB simulation}

\subsubsection{Running a simulation}

Building on the introduction of \pkg{contextual}'s core classes in the previous section, we can now put together the following five line MAB simulation to examine the 
performance of an $\epsilon$-greedy policy (covered in Section \ref{epsgreedy}) on a three-armed bandit:

\begin{CodeChunk}
\begin{CodeInput}
> library(contextual)
>
> bandit  <- ContextualBernoulliBandit$new(matrix(c(0.5, 0.2, 0.1), 1))
> policy  <- EpsilonGreedyPolicy$new(0.1)
> agent   <- Agent$new(policy,bandit)
> sim     <- Simulator$new(agent, simulations = 10000, horizon = 100)
> history <- sim$run()
\end{CodeInput}
\end{CodeChunk}

In these lines we start out by instantiating the \code{Bandit} subclass \code{ContextualBernoulliBandit} (covered in Section \ref{extending}) as \code{bandit}, with three 
Bernoulli arms, each offering a reward of one with reward probability $\theta$, and otherwise a reward of zero. For the current simulation, we have set the \code{bandit} arm 
probabilities of reward to respectively 0.5, 0.2 and 0.1 through the policy's \code{weights} parameter. In ContextualBernoulliBandit, the number of bandit arms equals number 
of columns of weight matrix. That is, the bandit instance's number of arms \code{bandit$k} now equals \code{3}, and, for this context-free setting, its number of feature 
dimension \code{bandit$d} remains \code{NULL}.

Next, we instantiate the \code{Policy} subclass \code{EpsilonGreedyPolicy} as object \code{policy}, with its \code{epsilon} parameter set to \code{0.1}.

We then assign both our \code{bandit} and our \code{policy} to \code{Agent} instance \code{agent}. This \code{agent} is added to a \code{Simulator} that is set to ten 
thousand \code{simulations}, each with a \code{horizon} of one hundred---that is, \code{simulator} runs ten thousand \code{simulations}\footnote{Each starting with a 
deterministically set random seed.} for one hundred time steps \code{t}.

In anticipation of Section \ref{classstructure}: For a policy to be able to initialize its parameters, during initialisation, an agent instance makes \code{bandit$k} (number 
of arms) and \code{bandit$d} (number of dimensions) available to its policy through calls to \code{policy$set_parameters(context_initial_params)} and 
\code{policy$initialize_theta(context_initial_params$k)}. On starting the simulation, the (potentially changing) number of arms and features dimension remain available for 
each time step \code{t} through respectively \code{context$k} and \code{context$d}.

Running the \code{Simulator} instance then starts several (by default, the number of CPU cores minus one) worker processes, dividing simulations over all parallel worker. 
For each simulation, for every time step \code{t}, \code{agents} runs through each of the four function calls that constitute their main loop.

A main loop that relates one on one to the four steps defined in our CMAB formalization from Section \ref{formalization}:

\begin{enumerate}
         \item[1)] \code{agent} calls \code{bandit$get_context(t)}. The \code{bandit} returns a named list that contains the current \code{d} $\times$ \code{k} dimensional 
	 feature matrix \code{context$X}, the number of arms \code{context$k} and the number of features per arm \code{context$d}.
         \item[2)] \code{agent} calls \code{policy$get_action(t, context)}. The \code{policy} computes which arm to play based on the current values in named lists 
	 \code{theta} and \code{context}. The \code{policy} returns a named list containing \code{action$choice}, which holds the index of the arm to play.
         \item[3)] \code{agent} calls \code{bandit$get_reward(t, context, action)}. The \code{bandit} returns a named list containing the \code{reward} for the \code{action} 
	 chosen in [2] and, optionally, an \code{optimal_reward}---when computable.
         \item[4)] \code{agent} calls \code{policy$set_reward(t, context, action, reward)}. The \code{policy} uses the \code{action} taken, the \code{reward} received, and 
	 the current \code{context} to update its set of parameter values in \code{theta}.
\end{enumerate}

\subsubsection{Results of the simulation}

On completion of all of its agents' simulation runs, the \code{Simulator} instance returns a \code{History} instance containing a complete log of all interactions. This 
history log can then, for example, be summarized and plotted:

\begin{CodeChunk}
\begin{CodeInput}
> summary(history)
\end{CodeInput}
\begin{CodeOutput}
Agents:

  EpsilonGreedy

Cumulative regret:

         agent   t  sims cum_regret cum_regret_var cum_regret_sd
 EpsilonGreedy 100 10000      9.115       101.4203      10.07077


Cumulative reward:

         agent   t  sims cum_reward cum_reward_var cum_reward_sd
 EpsilonGreedy 100 10000     40.816       119.8215       10.9463


Cumulative reward rate:

         agent   t  sims cur_reward cur_reward_var cur_reward_sd
 EpsilonGreedy 100 10000    0.40816       1.198215      0.109463

\end{CodeOutput}
\end{CodeChunk}

\begin{CodeChunk}
\begin{CodeInput}
> plot(history, type = "arms")
\end{CodeInput}
\end{CodeChunk}
\begin{figure}[H]
\centering
\includegraphics[width=.99\textwidth]{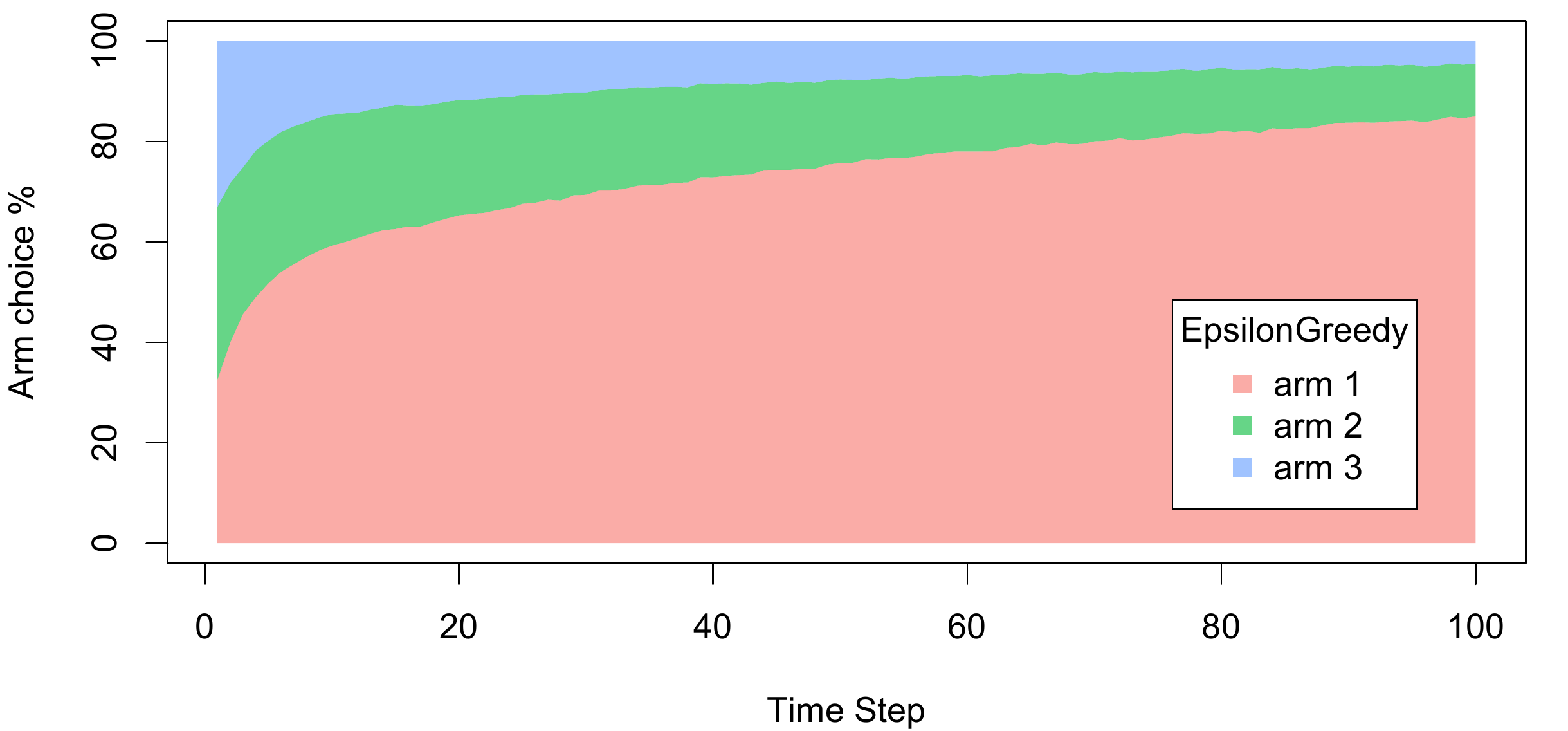}
\caption{The percentage each of the three arms of the bandit was chosen by the policy over all simulations per time step $t$. The plot represents the progression of this 
percentage for each time step from 1 to 100.}
\label{fig:section_2_3}
\end{figure}

\section{Basic usage} \label{basicusage}

The current section offers an overview of \pkg{contextual}'s predefined bandits and policies and further demonstrates how to run them.

\subsection{Implemented policies and bandits} \label{implbp}

Though contextual was designed to make it easy to develop custom bandit and policy classes, it is also possible to run basic simulations with just its built-in bandits and 
policies. See Table \ref{table:overview_policies} for an overview of all available policies and Table \ref{table:overview_bandits} for an overview of all bandits implemented 
up untill december 2018---where possible referencing their original papers.

\begin{table}[H]

\begin{tabularx}{\textwidth}{@{}lllllX@{}}
\toprule
\parnoteclear % tabularx will otherwise add each note thrice
& \textbf{$\epsilon$-Greedy} & \textbf{UCB} & \textbf{Thomspon Sampling} & \textbf{Other} & \textbf{Special} \\ \midrule
MAB & \begin{tabular}[c]{@{}l@{}}$\epsilon$-Greedy\parnote{\cite{Sutton1998e}}\\ $\epsilon$-First \end{tabular} & \begin{tabular}[c]{@{}l@{}}UCB1\parnote{\cite{Auer2002}} \\ 
UCB-tuned\parnote{\cite{Auer2002}} \end{tabular} & \begin{tabular}[c]{@{}l@{}}Thompson Sampling\parnote{\cite{Agrawal2011}} \\ BootstrapTS\parnote{\cite{Eckles2014}} 
\end{tabular} & \begin{tabular}[c]{@{}l@{}}Softmax\parnote{\cite{Vermorel2005}}\\ Gittins\parnote{\cite{Brezzi2002}}\end{tabular} & 
\multirow{2}{*}{\begin{tabular}[c]{@{}l@{}}Random\\ Oracle\\ LiF\parnote{\cite{Kaptein2016a}}\end{tabular}} \\ \cmidrule(r){1-5}
CMAB & Epoch-Greedy\parnote{\cite{Langford2008}} & \begin{tabular}[c]{@{}l@{}}LinUCB\parnote{\cite{Li2010}} \\ \end{tabular} & 
\begin{tabular}[c]{@{}l@{}}LinTS\parnote{\cite{Agrawal2012a}} \\ LogitBTS\parnote{\cite{Eckles2014}} \end{tabular} & & \\ \bottomrule
\end{tabularx}
\captionsetup{singlelinecheck = false, justification=justified}
\caption{An overview of \pkg{contextual}'s predefined contextual and context-free policy classes.}
\parnotes
\parnotereset
\label{table:overview_policies}
\end{table}

\begin{table}[H]
\begin{tabularx}{\textwidth}{@{}lllll@{}}
\toprule
\textbf{MAB} & \textbf{CMAB} & \textbf{Offline} & \textbf{Continuous} \\ \midrule
\begin{tabular}[t]{@{}l@{}}BasicBernoulliBandit\\ BasicGaussianBandit\end{tabular}  & \begin{tabular}[t]{@{}l@{}}ContextualBernoulli\\ ContextualLogit\\ ContextualHybrid\\ 
ContextualLinear\\ContextualWheel\parnote{\cite{Riquelme2018}}\end{tabular} & \begin{tabular}[t]{@{}l@{}}ReplayEvaluator\parnote{\cite{Li2011}}\\Bootstrap 
Replay\parnote{\cite{Nicol2014}}\\PropensityWeighting\textsuperscript{4}\\Direct Method\textsuperscript{4}\\ Doubly Robust\parnote{\cite{Dudik2011}}\end{tabular} & 
ContinuumBandit \\ \bottomrule
\end{tabularx}
\captionsetup{singlelinecheck = false, justification=justified}
\caption{An overview of \pkg{contextual}'s predefined synthetic and offline bandit classes.}
\parnotes
\label{table:overview_bandits}
\end{table}

\subsection{Running basic simulations} \label{basicsc}

In the current subsection we demonstrate how to run simulations with \pkg{contextual}'s predefined \code{Bandit} and \code{Policy} subclasses on the basis of a familiar 
bandit scenario.

\subsubsection{The scenario} \label{scen}

Since online advertising is one of the areas where bandit policies have found widespread application, we will use it as the setting for our basic bandit example. Generally, 
the goal in online advertising is to determine which out of several ads to serve a visitor to a particular web page. Translated to a bandit setting, in online advertising:

\begin{itemize}
         \item The context is usually determined by visitor and web page characteristics.
         \item Arms are represented by the pool of available ads.
         \item An action equals a shown ad.
         \item Rewards are determined by a visitor clicking (a reward of 1) or not clicking (a reward of 0) on the shown ad.
\end{itemize}

For the current example, we limit the number of advertisements we want to evaluate to three, and set ourselves the objective of finding which policy would offer us the 
highest total click-through rate\footnote{Click-through rate (CTR) is the ratio of users who click on a specific link or ad to the number of total users who view it 
\citep{Briggs1997}.} over four hundred impressions.

\subsubsection{Evaluating context-free policies} \label{ncp}

Before we are able to evaluate any policies, we first need to model our three ads---each with a different probability of generating a click---as the arms of a bandit. For 
our current simulation we choose to model the ads with the weight-based \code{ContextualBernoulliBandit}, as this allows us to set weights determining the average reward 
probability of each arm. As we are comparing context-free bandits, here, we specify only one row in \code{ContextualBernoulliBandit}'s weight-matrix, representing average 
reward probabilities or theta's per arm (in the following section we will add rows to the matrix, each additional row representing one context feature's weight per arm). As 
can be observed in the source code below, for the current simulation, we set the weights of the arms to respectively $\theta_1 = 0.8$, $\theta_2  = 0.4$ and $\theta_3 = 
0.2$.

We also choose two context-free policies to evaluate and compare:

\begin{itemize}
         \item \code{EpsilonFirstPolicy}: explores the three ads uniformly at random for a preset period and from thereon exploits the ad with the best click-through 
	 rate\footnote{A type of policy also known as an A/B test \citep{Kohavi2007}.}. For our current scenario, we set the exploration period to one hundred impressions. A 
	 formal definition and implementation of the algorithm can be found in Section \ref{epsfirst}.

         \item \code{EpsilonGreedyPolicy}: explores one of the ads uniformly at random $\epsilon$ of the time and exploits the ad with the best current click-through rate $1 
	 - \epsilon$ of the time. For our current scenario, we set $\epsilon = 0.4$. For a formal definition and implementation see Section \ref{epsgreedy}.
\end{itemize}

Next, we assign the \code{bandit} and our two \code{policy} instances to two \code{agents}. Finally, we assign a \code{list} holding both \code{agents} to a \code{Simulator} 
instance, set the \code{simulator}'s horizon to four hundred and the number of repeats to ten thousand, run the simulation, and \code{plot()} its results:

\begin{Code}
# Load and attach the contextual package.
library(contextual)
# Define for how long the simulation will run.
horizon <- 400
# Define how many times to repeat the simulation.
simulations <- 10000
# Define the probability that each ad will be clicked.
click_probabilities <- matrix(c(0.6, 0.4, 0.2), nrow = 1, ncol = 3, byrow = TRUE)
# Initialize a SyntheticBandit, which takes probabilites per arm for an argument.
bandit <- ContextualBernoulliBandit$new(weights = click_probabilities)
# Initialize EpsilonGreedyPolicy with a 40 percent exploiration rate.
eg_policy <- EpsilonGreedyPolicy$new(epsilon = 0.4)
# Initialize EpsilonFirstPolicy with a .25 x 400 = 100 step exploration period.
ef_policy <- EpsilonFirstPolicy$new(epsilon = 0.25, N = horizon)
# Initialize two Agents, binding each policy to a bandit.
ef_agent <- Agent$new(ef_policy, bandit)
eg_agent <- Agent$new(eg_policy, bandit)
# Assign both agents to a list.
agents <- list(ef_agent, eg_agent)
# Initialize a Simulator with the agent list, horizon, and number of simulations.
simulator <- Simulator$new(agents, horizon, simulations, do_parallel = TRUE)
# Now run the simulator.
history <- simulator$run()
# Finally, plot the average reward per time step t
plot(history, type = "average", regret = FALSE)
# And the cumulative reward rate, which equals the Click Through Rate)
plot(history, type = "cumulative", regret = FALSE, rate = TRUE)

\end{Code}
\begin{figure}[H]
\centering
\includegraphics[width=.99\textwidth]{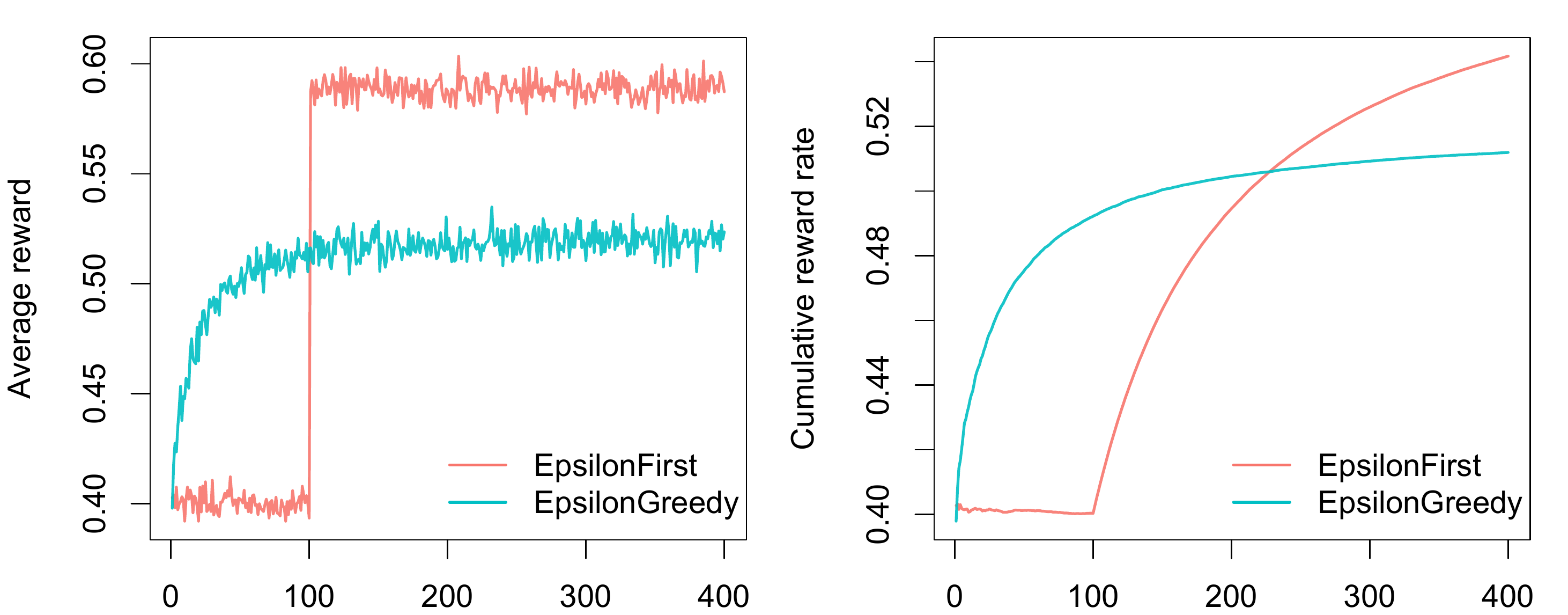}
\caption{Average reward (left) and cumulative reward rate (equaling click-through rate, right) of $\epsilon$-first and $\epsilon$-greedy policies.}
\label{fig:section_3_2_1}
\end{figure}

As can be observed in Figure \ref{fig:section_3_2_1}, within our horizon of $T = 400$, \code{EpsilonFirstPolicy} has accumulated more rewards than 
\code{EpsilonGreedytPolicy}. It is easy to see why: The winning arm is better than the other two---by a margin. So \code{EpsilonFirstPolicy} has no difficulty in finding the 
optimal arm within its exploration period of one hundred impressions. Up to that point, \code{EpsilonGreedyPolicy} had the advantage of a headstart, as it was already able 
to exploit for $1- \epsilon$ or sixty percent of the time. But from one hundred impressions on, \code{EpsilonFirstPolicy} switches from full exploration to full exploitation 
mode. In contrast to \code{EpsilonGreedyPolicy}, it is now able to exploit the arm that proved best during exploration all of the time. As a result, it catch up with (and 
then surpass) the rewards accumulated by \code{EpsilonGreedyPolicy} within less than one hundred and fifty impressions.

\subsubsection{Adding context} \label{addingctx}

If that is all we know of our visitors, we expect the results to be stationary over time, and these are the only policies available, the choice is clear: for this scenario, 
you would pick \code{EpsilonFirstPolicy}\footnote{Also: if our bandit represents our visitors' click behavior realistically, if our policies' parameters are optimal, 
etcetera.}. However, if we have contextual information on our visitors---for instance, their age---we might be able to do better. Let us suggest that we expect that some of 
our ads are more effective for older visitors, and other ads more effective for younger visitors.

To incorporate this expectation in our simulation, we need to change the way our bandit generates its rewards. Fortunately, in the case of our 
\code{ContextualBernoulliBandit}, the introduction of two contextual features only requires the addition of a single row to its weight matrix---as 
\code{ContextualBernoulliBandit} parses each of the $d$ rows of its weight matrix as a binary contextual feature randomly selected or sampled $1/d$ of the time.

We took care to set the combined weight per arm such that each arm generates the same average rewards as previously. So we do not expect a substantial difference with the 
last simulation's outcome for our context-free policies \code{EpsilonFirstPolicy} and \code{EpsilonGreedyPolicy}.

We therefore now also include the contextual \code{LinUCBDisjointPolicy} \citep{Li2010}, which, in assuming its reward function is a linear function of the context, should 
be able to incorporate our new contextual information into its decision-making process. See \ref{linucbc} for a detailed description and implementation details of this 
policy. Now let us rerun the simulation:

\begin{Code}

#                        +-----+----+----------->  arms:  three ads
#                        |     |    |
click_probs <- matrix(c(0.5,  0.7, 0.1,  # -> context 1: older (p=.5)
                        0.7,  0.1, 0.3), # -> context 2: young (p=.5)

                        nrow = 2, ncol = 3, byrow = TRUE)

# Initialize a SyntheticBandit with contextual weights
context_bandit <- ContextualBernoulliBandit$new(weights = click_probs)
# Initialize LinUCBDisjointPolicy
lucb_policy    <- LinUCBDisjointPolicy$new(0.6)
# Initialize three Agents, binding each policy to a bandit.
ef_agent       <- Agent$new(ef_policy,   context_bandit)
eg_agent       <- Agent$new(eg_policy,   context_bandit)
lucb_agent     <- Agent$new(lucb_policy, context_bandit)
# Assign all agents to a list.
agents <- list(ef_agent, eg_agent, lucb_agent)
# Initialize a Simulator with the agent list, horizon, and nr of simulations
simulator <- Simulator$new(agents, horizon, simulations)
# Now run the simulator.
history <- simulator$run()
# Finally, plot the average reward..
plot(history, type = "average", regret = FALSE)
# And the cumulative reward rate again.
plot(history, type = "cumulative", regret = FALSE, rate = TRUE)
\end{Code}
\begin{figure}[H]
\centering
\includegraphics[width=.99\textwidth]{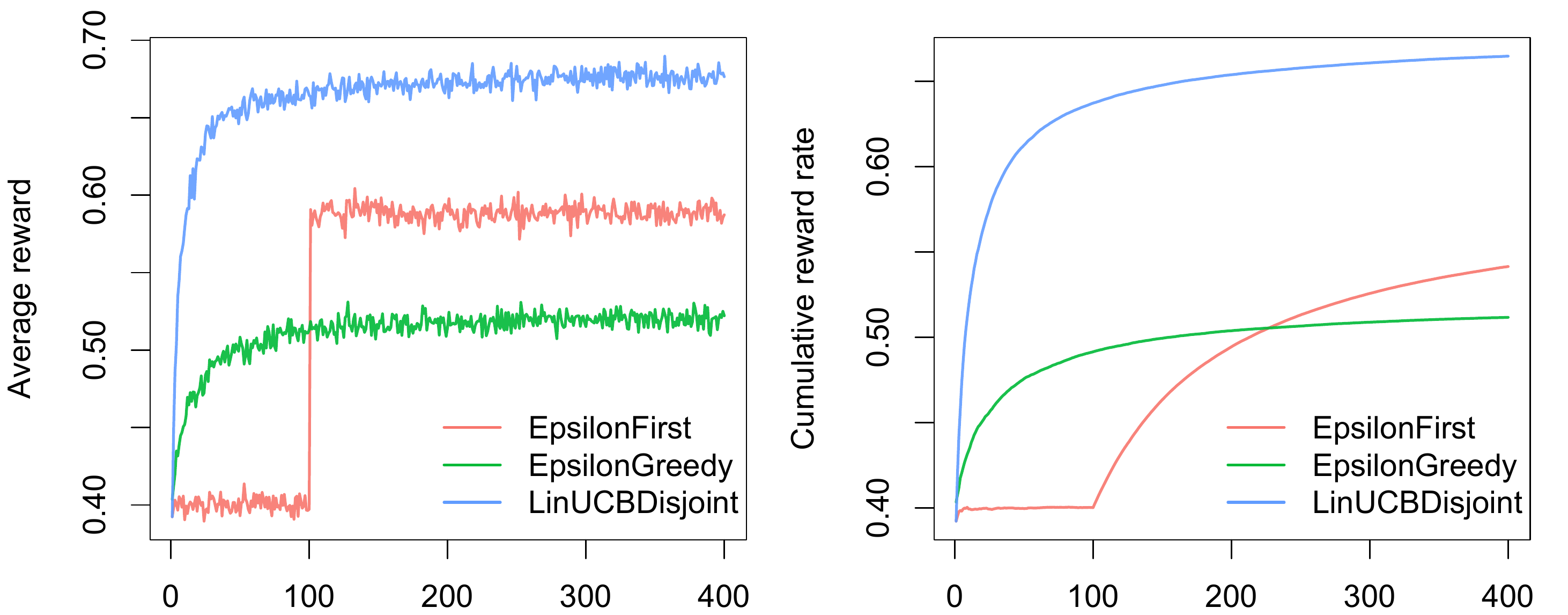}
\caption{Average reward (left) and cumulative reward rate (equaling click-through rate, right) of LinUCB, $\epsilon$-first and $\epsilon$-greedy policies.}
\label{fig:section_3_2_2}
\end{figure}

As can be observed in Figure \ref{fig:section_3_2_2}, both context-free bandit's results do indeed not do better than before. On the other hand, \code{LinUCBDisjointPolicy} 
does very well, as it indeed proves able to map its rewards to the available contextual features.

Of course, the simulations in the current section are not very realistic. One way of creating more realistic simulations would be to write a Bandit subclass with a more 
complex generative model. Section \ref{subclpb} shows how to get started with that. Another option would be to evaluate policies on an offline dataset---more on how to go 
about that in Section \ref{offl}.

\section{Core classes} \label{classstructure}

The current section offers additional background information on \pkg{contextual}'s class structure---both on the R6 class system \cite{R6} and on each of the six previously 
introduced core \pkg{contextual} classes. Together with the information in the next section, on bandit and policy implementation, this should be able to get you up and 
running with developing your own custom \code{Bandit} and \code{Policy} subclasses.

\subsection{Choice for the R6 class system} \label{classsystem}

Though widely used as a procedural language, \proglang{R} offers several Object Oriented (OO) systems, which can significantly help in structuring the development of more 
complex packages. Out of the OO systems available (S3, S4, R5 and R6), we settled on R6, as it offered several advantages compared to the other options. Firstly, it 
implements a mature object-oriented\footnote{In object-oriented programming, the developer compartmentalizes data into objects, whose behavior and contents are described 
through the declaration of classes. Its benefits include reusability, refactoring, extensibility, ease of maintenance and efficiency. See, for instance, 
\cite{Wirfs-Brock1990} for a general introduction to the princples of Object Oriented software design, and \cite{wickham2014advanced} for more information of the use of OOP 
in \proglang{R}.} design when compared to S3. Secondly, its classes can be accessed and modified by reference---which offers the added advantage that R6 classes are 
instantly recognizable for developers with a background in programming languages such as \proglang{Java} or \proglang{C++}. Finally, when compared to the older R5 reference 
class system, R6 classes are much lighter-weight and, as they do not make use of S4 classes, do not require the \pkg{methods} package.

\subsection{Main classes} \label{mainclasses}

In this section, we go over each of \pkg{contextual}'s six main classes in some more detail---with an emphasis on the \code{Bandit} and \code{Policy} classes. To clarify 
\pkg{contextual}'s class structure, we also include two UML diagrams (UML, or "unified modeling language" presents a standardized way to visualize the overall class 
structure and general design of a software application or framework \citep{Rumbaugh2004}). The UML class diagram shown in Figure \ref{fig:contextual_class} on page 
\pageref{fig:contextual_class} visualizes \pkg{contextual}'s static object model, showing how its classes inherit from, and interface with, each other. The UML sequence 
diagram in figure Figure \ref{fig:contextual_sequence} on page \pageref{fig:contextual_sequence}, on the other hand, illustrates how \pkg{contextual}'s classes interact 
dynamically over time.

\subsubsection{Simulator}

\includegraphics[width=\textwidth]{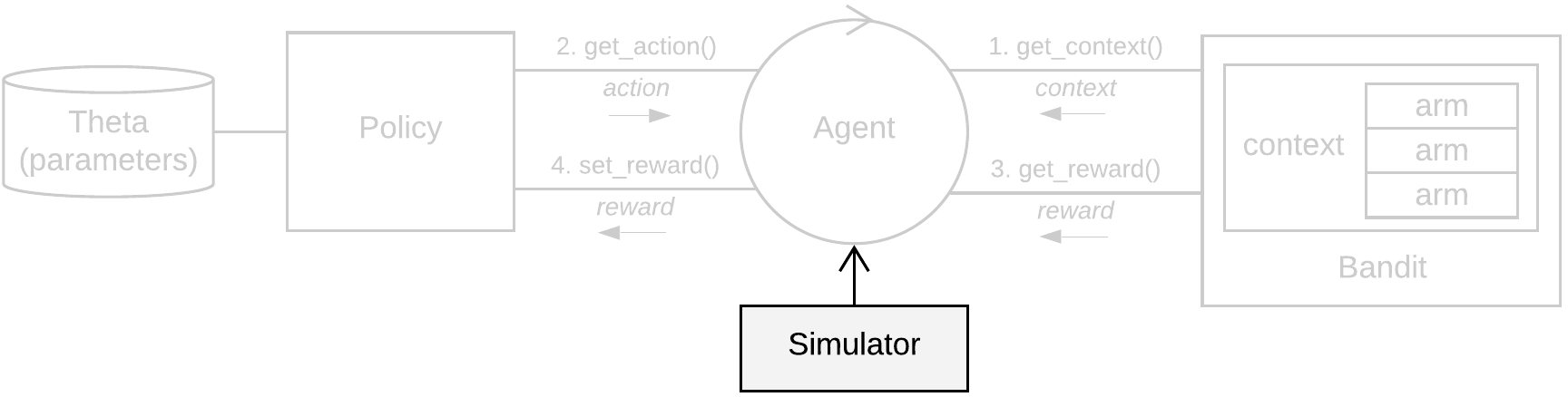}

A \code{Simulator} instance is the entry point of any \pkg{contextual} simulation. It encapsulates one or more \code{Agents}, clones them if necessary, runs the 
\code{Agents} (in parallel, by default), and saves the log of all of the \code{Agents} interactions to a \code{History} object:

\begin{Code}
history <- Simulator$new(agents = agent, horizon = 10, simulations = 10)$run()
\end{Code}

By default, for performance reasons, a \code{Simulator} does not save \code{context} matrices and the (potentially deeply nested) \code{theta} list to its \code{History 
log}---though this can be changed  by setting either \code{save_context} and \code{save_theta} arguments set to \code{TRUE}.

To specify how to run a simulation and which data is to be saved to a \code{Simulator} instance's \code{History} log, a \code{Simulator} object can be configured through, 
among others, the following arguments:

\begin{itemize}
   \item{\code{agents}}{
     \code{[NULL]} An \code{Agent} instance, or a \code{list} of \code{Agent} instances to be run by the instantiated \code{Simulator}.
   }
   \item{\code{horizon}}{
      \code{[100]} The T time steps to run the instantiated \code{Simulator}.
   }
   \item{\code{simulations}}{
      \code{[100]} How many times to repeat each agent's simulation with a new seed on each repeat (itself deterministically derived from set\_seed).
   }
   \item{\code{save_context}}{
      \code{[FALSE]} Save context matrices \code{X} to the \code{History} log during a simulation?
   }
   \item{\code{save_theta}}{
     \code{[FALSE]}  Save the parameter list \code{theta} to the \code{History} log during a simulation?
   }
   \item{\code{do_parallel}}{
     \code{[TRUE]}  Run \code{Simulator} processes in parallel?
   }
   \item{\code{worker_max}}{
      \code{[NULL]}  Specifies how many parallel workers are to be used, when \code{do_parallel} is \code{TRUE}. If unspecified, the amount of workers defaults to 
      \code{max(workers_available)-1}.
   }
   \item{\code{set_seed}}{
      \code{[0]}  Sets the seed of \proglang{R}'s random number generator for the current \code{Simulator}.
   }
   \item{\code{progress_file}}{
       \code{[FALSE]}  If \code{TRUE}, \code{Simulator} writes \code{progress.log} and \code{doparallel.log}
       files to the current working directory, allowing you to keep track of \code{workers}, iterations,
       and potential errors when running a \code{Simulator} in parallel.
   }
   \item{\code{include_packages}}{
       \code{[NULL]}  List of packages that (one of) the policies depend on. If a \code{Policy} requires an
       \proglang{R} package to be loaded, this option can be used to load that package on each of the workers.
       Ignored if \code{do_parallel} is \code{FALSE}.
   }
   \item{\code{reindex}}{
      \code{[FALSE]} If \code{TRUE}, removes empty rows from the \code{History} log,
      re-indexes the \code{t} column, and truncates the resulting data to the shortest simulation
      grouped by agent and simulation.
   }
\end{itemize}

The \code{Simulator} class has been designed to make it as easy as possible to replace its default parallel backend for other \pkg{foreach} type backends. In the 
\code{alt_par_backend_examples} subdirectory of \pkg{contextual}'s root \code{demo} directory we offer working examples of \pkg{doAzureParallel}, \pkg{doRedis} and 
\pkg{doMPI} \code{Simulator} subclasses. As a result of \pkg{contextual}'s object oriented structure, such Simulator subclasses can be implemented in just a few lines of 
code, for instance in the case of the \pkg{doMPI} subclass:

\begin{Code}
MPISimulator <- R6::R6Class(
  inherit = Simulator,
  public = list(
    # Register a foreach parallel backend.
    register_parallel_backend = function() {
      super$cl <- doMPI::startMPIcluster()
      doMPI::registerDoMPI(super$cl)
      # Also make sure to set Simulator's super$workers.
      super$workers = foreach::getDoParWorkers()
      message(paste0("MPI workers: ", super$workers))
    },
    # If necessary, clean up.
    stop_parallel_backend = function() {
      try({
        doMPI::closeCluster(super$cl)
      })
    }
  )
)
\end{Code}

\subsubsection{Agent}

\includegraphics[width=\textwidth]{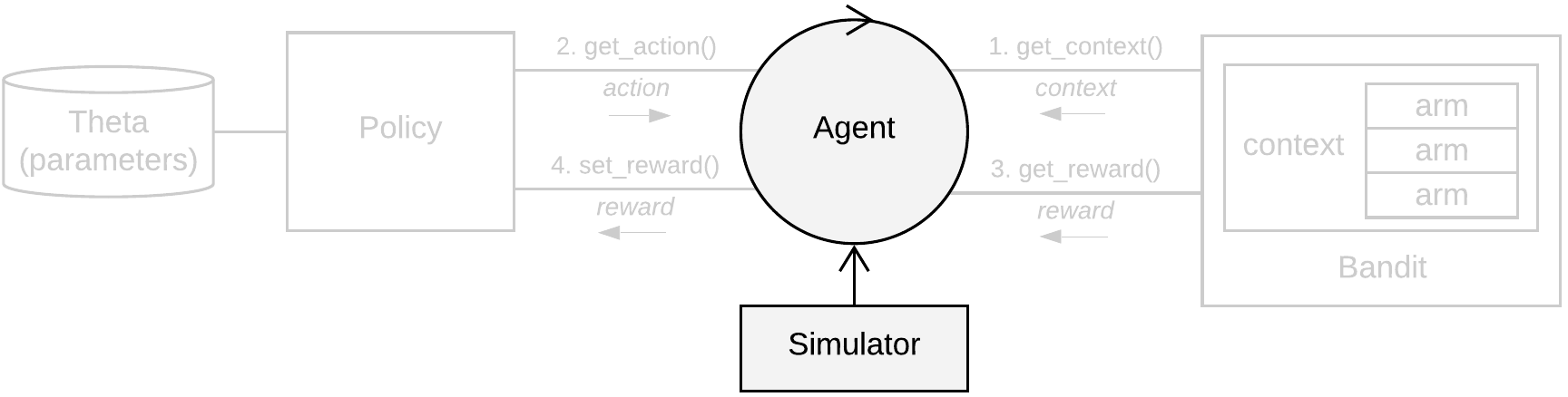}

To ease the encapsulation of parallel \code{Bandit} and \code{Policy} simulations, \code{Agent} is responsible for the flow of information between and the running of one 
\code{Bandit} and \code{Policy} pair, for example:

\begin{Code}
policy             <- EpsilonGreedyPolicy$new(epsilon = 0.1)
bandit             <- ContextualBernoulliBandit$new(weights = c(0.9, 0.1, 0.1))
agent              <- Agent$new(policy,bandit, name = "EG Agent One")
\end{Code}

It keeps track of agent and policy related timesteps \code{t} and makes sure that, at each agent time step \code{agent_t}, the four main \code{Bandit} and \code{Policy} CMAB 
methods are called in correct order, one after the other:

\begin{Code}
# Abstracted skeleton code representation of Agent class
Agent <- R6::R6Class(
  public = list(
    agent_t = 0,
    policy_t = 0,
    #...
    do_step = function() {
      agent_t  <<- agent_t + 1L
      context   <- bandit$get_context(agent_t)
      action    <- policy$get_action (policy_t, context)
      reward    <- bandit$get_reward (agent_t, context, action)
      if (is.null(reward)) {
        theta   <- NULL
      } else {
        theta   <- policy$set_reward(policy_t, context, action, reward)
        policy_t  <<- policy_t + 1L
      }
      list(context, action, reward, theta, policy_t)
    }
    #...
  )
)
\end{Code}

Its main function is \code{do_step()}, generally called by a \code{Simulator} object (or, more specifically, by the \code{Simulator}-started parallel worker that is 
repsonsible for this particular \code{Agent}):

\begin{itemize}
   \item{\code{do_step()}}{
      Completes one time step \code{t} by consecutively calling
      \code{bandit$get_context()}, \code{policy$get_action()}, \code{bandit$get_reward()} and \code{policy$set_reward()}.
    }
\end{itemize}

\subsubsection{Bandit}

In \pkg{contextual}, any bandit implementation is expected to subclass and extend the \code{Bandit} superclass. It is then up to these subclasses themselves to provide an 
implementation for each of its abstract methods.

\code{Bandits} are responsible for the generation of (either synthetic or offline) contexts and rewards. On initialisation, a \code{Bandit} subclass has to define the number 
of arms \code{self$k} and the number of contextual feature dimensions \code{self$d}. For each \emph{t} = \{1, \ldots, T\} a \code{Bandit} then generates a \code{list} 
containing current context as either a \code{d} dimensional context vector, or a \code{d} $\times$ \code{k} dimensional matrix\footnote{For each $t$, a \code{Bandit} 
generated matrix describes both features shared by all arms and features that differ per arm. In other words, each column of this matrix represents a single arm's feature 
vector---combining, for instance, overall user and arm specific article weights. See also Section \ref{formalization}.} \code{context$X}, the number of arms in 
\code{context$k} and the number of features in \code{context$d} (Note: in context-free scenario's, \code{context$X} can be omitted):

\includegraphics[width=\textwidth]{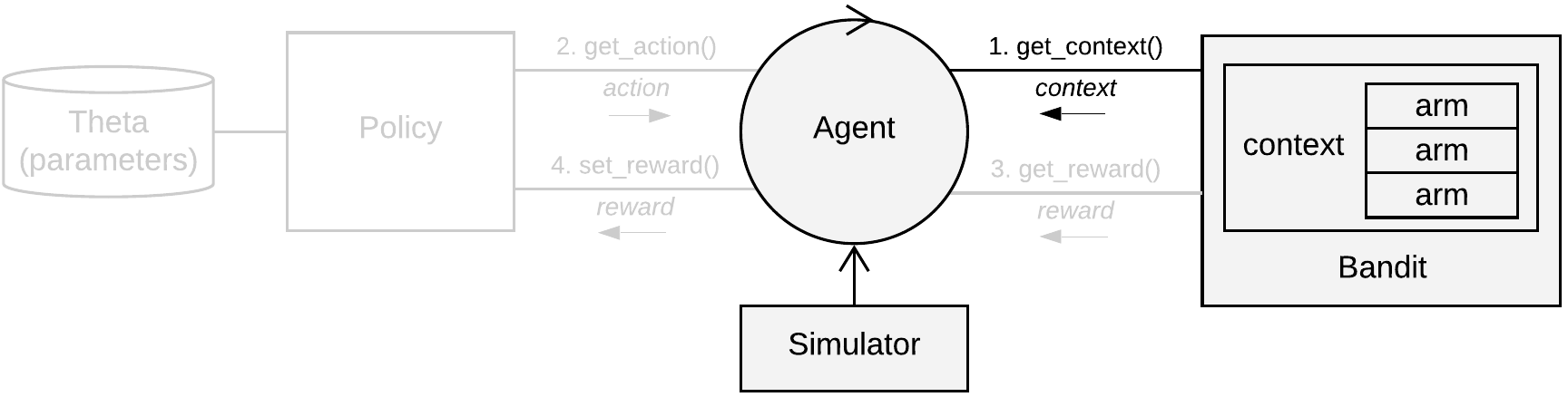}

On receiving the index of a \code{Policy}-chosen arm through \code{action$choice}, \code{Bandit} is expected to return a named \code{list} containing at least 
\code{reward$reward} and, where computable, \code{reward$optimal}:

\includegraphics[width=\textwidth]{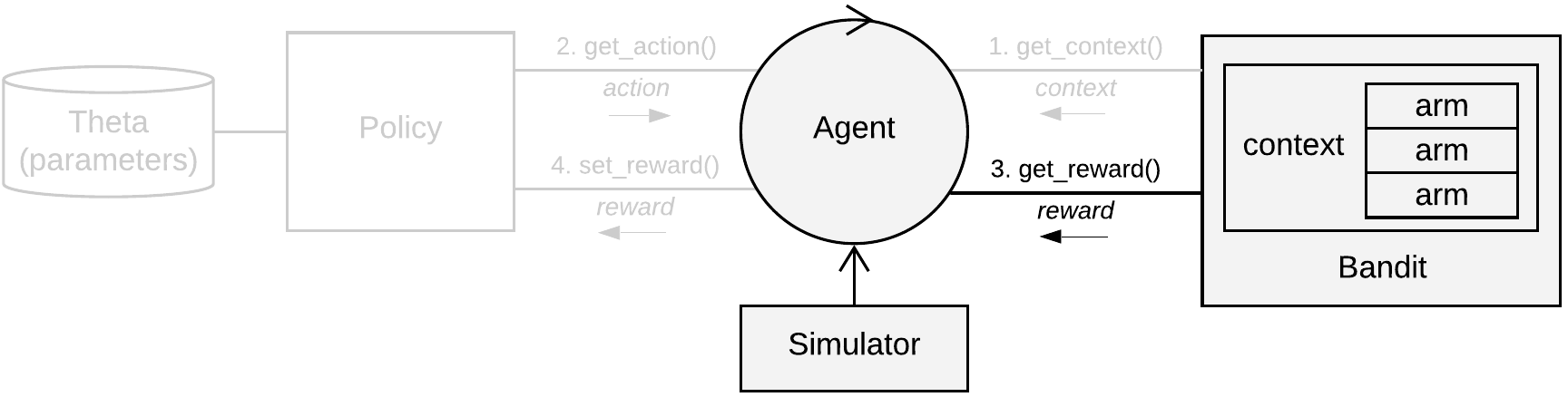}

\pagebreak

The following skeleton code gives an overview of how the above is implemented in \pkg{contextual}'s \code{Bandit} superclass:

\begin{Code}
Bandit <- R6::R6Class(
  class    = FALSE,
  public   = list(
    k           = NULL,  # Number of arms (integer, required)
    d           = NULL,  # Dimensions of context vector (integer, required)
    ..
    class_name  = "Bandit", # Bandit name - required (character)
    initialize  = function() {
      # Initialize Bandit. Set self$d and self$k here.
    },
    get_context = function(t) {
      stop("Bandit subclass needs to implement get_context()", call. = FALSE)
      # Return a list with number of arms self$k, number of feature dimensions
      # self$d and, where applicable, a self$d dimensional context vector or 
      # self$d x self$k dimensional context matrix X.
      list(X = context, k = arms, d = features) # nocov
    },
    get_reward = function(t, context, action) {
      stop("Bandit subclass needs to implement get_reward()", call. = FALSE)
      # Return a list with the reward of the chosen arm and, if available, 
      # optimal arm reward and index
      list(reward = reward_for_choice_made, optimal_reward = optimal_reward, optimal_arm = optimal_arm) # nocov
    },
    ...
  )
)
\end{Code}

The main \code{Bandit} functions can be futher detailed as follows:

\begin{itemize}
   \item{\code{new()}}{ generates and instantializes a new \code{Bandit} instance. }

   \item{\code{get_context(t)}}{
         \itemize{
          \item \code{t}: \code{[integer]}, time step \code{t}.
      }

      Returns a named \code{list}
      containing the current \code{d x k} dimensional matrix \code{context$X},
      the number of arms \code{context$k} and the number of features \code{context$d}.
  }

   \item{\code{get_reward(t, context, action)}}{
      \itemize{
          \item \code{t}: \code{[integer]}, time step \code{t}.
          \item \code{context}: \code{[list]}, containing the current \code{context$X} (d x k context matrix), \code{context$k} (number of arms) and \code{context$d} (number 
	  of context features) (as set by \code{bandit}).
          \item \code{action}:  \code{[list]}, containing \code{action$choice} (as set by \code{policy}).
      }

      Returns a named \code{list} containing \code{reward$reward} and, where computable,
         \code{reward$optimal} (used by "oracle" policies and to calculate regret).
  }

   \item{\code{post_initialization()}}{
      Called after class and seed initialisation, but before the start of the simulation.
      Set random values that remain available throughout the life of a \code{Bandit} here.
   }

   \item{\code{generate_bandit_data(n = horizon)}}{
      Called after class and seed initialisation, but before the start of a simulation.
      Only called when \code{bandit$precaching} is set to \code{TRUE} (default \code{FALSE}).
      Pregenerate $n$ \code{contexts} and \code{rewards} here.
   }
\end{itemize}

As already previously indicated in Table \ref{table:overview_bandits} in Section \ref{implbp} \code{Bandit}, \pkg{contextual} already contains several predefined Bandits, 
such as:

\begin{itemize}
         \item \code{BasicBernoulliBandit}: This basic (context-free) k-armed bandit synthetically generates rewards based on a weight vector.
         \item \code{BasicGaussianBandit}: Context-free Gaussian multi-armed bandit.
         \item \code{ContextualBernoulliBandit}: An example of a more complex and versatile synthetic bandit. It pregenerates both a randomized context matrix and reward 
	 vectors
         \item \code{ContextualLinearBandit}: Samples data from linearly parameterized arms.
         \item \code{ContextualWheelBandit}: The Wheel bandit game offers an artificial problem where the need for exploration is smoothly parameterized through an 
	 exploration parameter \citep{Riquelme2018}.
         \item \code{ContextualLogitBandit}: Samples data from a basic logistic regression model.
         \item \code{ContinuumBandit}: Bandit where arm(s) are chosen from a subset of the real line and mean rewards are assumed to be a continuous function of the arms.
         \item \code{OfflineReplayEvaluatorBandit}: Replays offline data to generate its contexts and rewards.  \cite{Li2010}.
\end{itemize}

Each of these bandits can either be run directly or serve as templates or superclasses for custom \code{Bandit} implementation(s).

\subsubsection{Policy}

\code{Policy} is another often subclassed contexual superclass. Just like the \code{Bandit} superclass, \code{Policy} is an abstract class that declares methods without 
itself offering an implementation. Any \code{Policy} subclass is therefore expected to implement \code{get_action()} and \code{set_reward()}. Also, any parameters that keep 
track or summarize \code{context}, \code{action} and \code{reward} values are required to be saved to \code{Policy}'s \textit{named list} \code{theta}.

On every \emph{t} = \{1, \ldots, T\}, a policy receives either a \code{d}-dimensional vector or a \code{d} $\times$ \code{k} dimensional matrix \code{context$X}. To make 
sure a policy supports both contextual feature vectors and matrices in \code{context$X}, it is suggested any contextual policy makes use of \pkg{contextual}'s 
\code{get_arm_context(context$X, arm)} utility function to obtain the current context for a particular arm, \code{and get_full_context(X, d, k)} where a policy needs to 
obtain of the full \code{d x k} matrix.

Next to the context vector or matrix, a policy also receives, at least, the current number of \code{Bandit} arms in \code{context$k}, and the current number of contextual 
features in \code{context$d}. It has to compute which of the \code{k} \code{Bandit} arms to pull by taking into account this contextual information plus the policy's current 
parameter values stored in the named list \code{theta}. On selecting an arm, the policy then returns its index as \code{action$choice}:

\includegraphics[width=\textwidth]{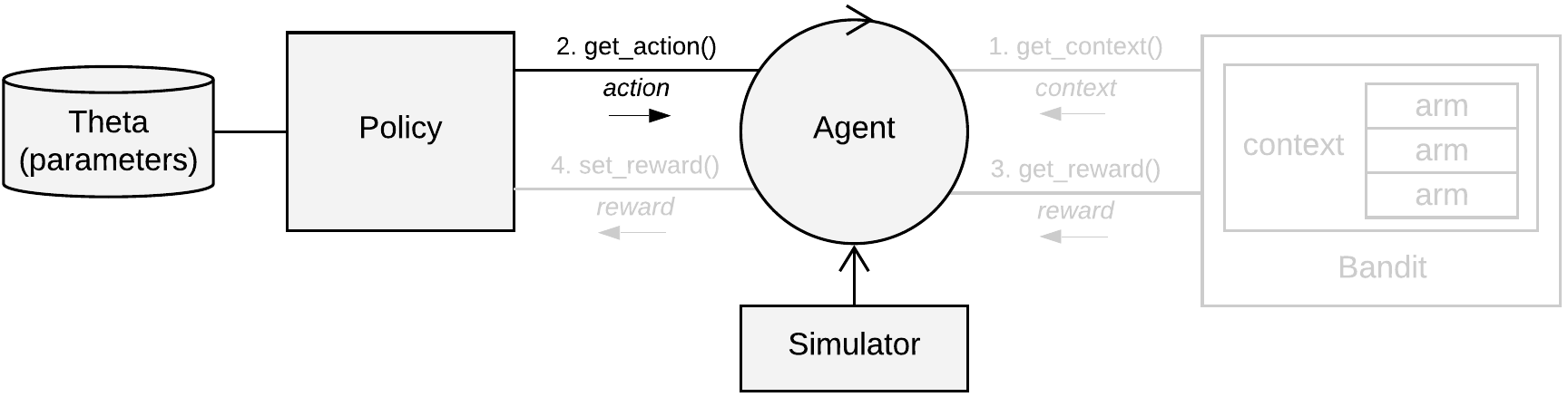}

On pulling a \code{Bandit} arm the policy receives a \code{Bandit} reward through \code{reward$reward}. In combination with the current \code{context$X} and 
\code{action$choice}, this reward can then be used to update to the policy's parameters as stored in list \code{theta}:

\includegraphics[width=\textwidth]{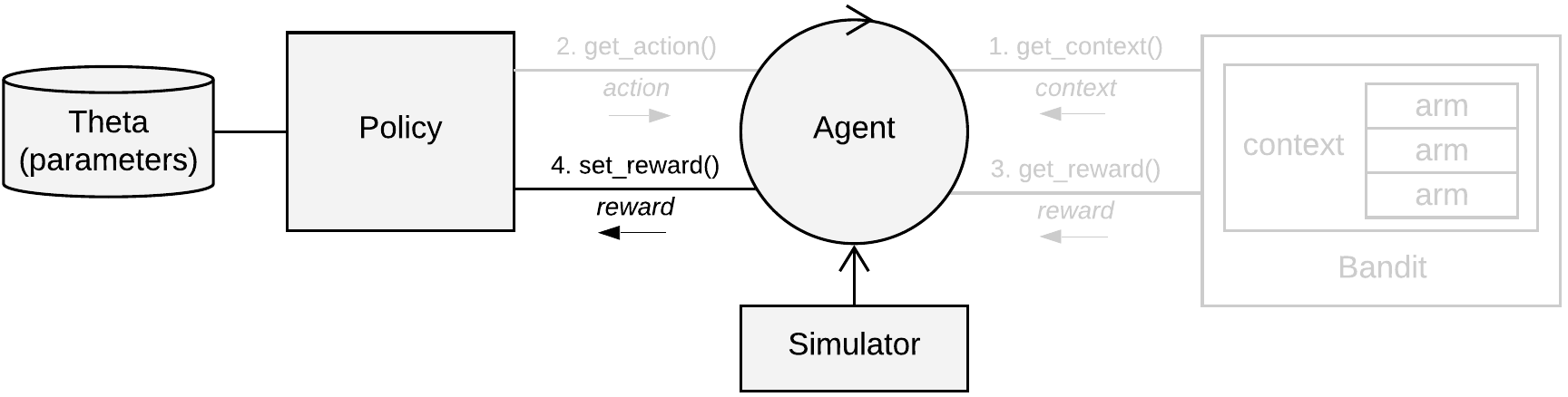}

Note: in context-free scenario's, \code{context$X} can be omitted.

The following skeleton code gives an overview of how the above is implemented in \pkg{contextual}'s \code{Policy} superclass:

\begin{Code}
Policy <- R6::R6Class(
  portable = FALSE,
  class = FALSE,
  public = list(
    action        = NULL,      # action results (list)
    theta         = NULL,      # policy parameters theta (list)
    theta_to_arms = NULL,      # theta to arms "helper" (list)
    class_name    = "Policy",  # policy name - required (character)
    initialize = function() {
      self$theta  <- list()    # initializes theta list
      self$action <- list()    # initializes action list
    },
    ...
    get_action = function(t, context) {
      # Selects arm based on theta & context, returns it in action$choice
      stop("Policy$get_action() has not been implemented.")
    },
    set_reward = function(t, context, action, reward) {
      # Updates parameters in theta based on reward awarded by bandit
      stop("Policy$set_reward() has not been implemented.")
    },
    ...
  )
)
\end{Code}

\code{Policy}'s main functions can be futher detailed as follows:

\begin{itemize}

   \item{\code{new()}}{
     Generates and initializes a new \code{Policy} object.
   }

   \item{\code{get_action(t, context)}}{
      \itemize{
          \item \code{t}: \code{[integer]}, time step \code{t}.
          \item \code{context}: \code{[list]}, containing the current \code{context$X} (d x k context matrix), \code{context$k} (number of arms) and \code{context$d} (number 
	  of context features)
      }.

      Computes which arm to play based on the current values in named list \code{theta} and the current \code{context}. \\Returns a named list containing 
      \code{action$choice}, which holds the index of the arm to play.
   }

   \item{\code{set_reward(t, context, action, reward)}}{
      \itemize{
          \item \code{t}: \code{[integer]}, time step \code{t}.
          \item \code{context}: \code{[list]}, containing the current \code{context$X} (d x k context matrix), \code{context$k} (number of arms) and \code{context$d} (number 
	  of context features).
          \item \code{action}:  \code{[list]}, containing \code{action$choice} (as set by \code{policy}).
          \item \code{reward}:  \code{[list]}, containing \code{reward$reward} and, if available, \code{reward$optimal} (as set by \code{bandit}).
      }

       Returns the set of updated parameters in list \code{theta}.
    }

   \item{\code{set_parameters()}}{
    Helper function, called during a Policy's initialisation, assigns the values it finds in list \code{self$theta_to_arms} to each of the Policy's k arms. The parameters 
    defined here can then be accessed by arm index in the following way: \code{theta[[index_of_arm]]$parameter_name}.
   }
\end{itemize}

\subsubsection{History}

\includegraphics[width=\textwidth]{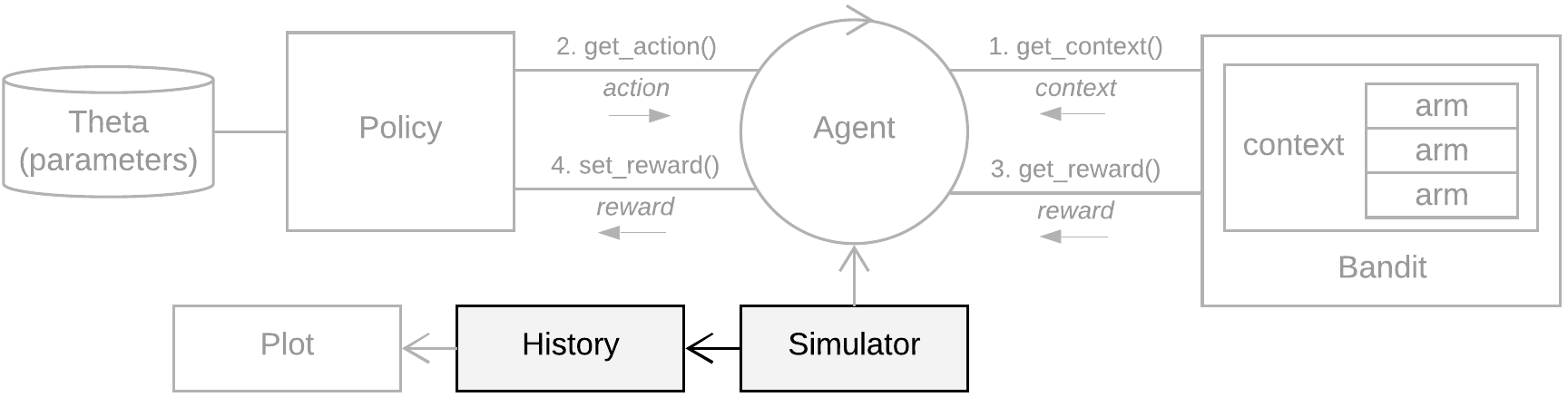}

A \code{Simulator} aggregates the data acquired during a simulation in a \code{History} object's private \code{data.table} log. It also calculates per agent average 
cumulative reward, and, when the optimal outcome per \code{t} is known, per agent average cumulative regret. It is furthermore possible to \code{plot()} a \code{History 
object}, \code{summarize()} it, or obtain, for example, a \code{data.frame()} or a \code{data.table()} from any \code{History} instance:

\begin{Code}
history             <- Simulator$new(agent)$run()
dt                  <- history$get_data_table()
df                  <- history$get_data_frame()
cumulative_regret   <- history$cumulative(regret = TRUE)
\end{Code}

Some other \code{History} functions:

\begin{itemize}
 \item{\code{set(index,
                  t,
                  action,
                  reward,
                  policy_name,
                  simulation_index,
                  context_value = NA,
                  theta_value = NA)}}{
    Stores one row of simulation data. Generally not called directly,
    but rather through a \code{Simulator} instance.
 }
 \item{\code{save(filename = NA)}}{
    Writes \code{History} to a file with name \code{filename}.
 }
 \item{\code{load(filename, interval = 0)}}{
    Reads a \code{History} log file with name \code{filename}.
    If \code{interval} is larger than 0, every other \code{interval} row of data is read instead of the
    full data file.
 }
 \item{\code{reindex(truncate = TRUE)}}{
    Removes empty rows from the \code{History} log, reindexes the \code{t} column, and,
    when \code{truncate} is \code{TRUE}, truncates the resulting data to the number of rows of the shortest
    simulation.
 }
\end{itemize}

\subsubsection{Plot}

\includegraphics[width=\textwidth]{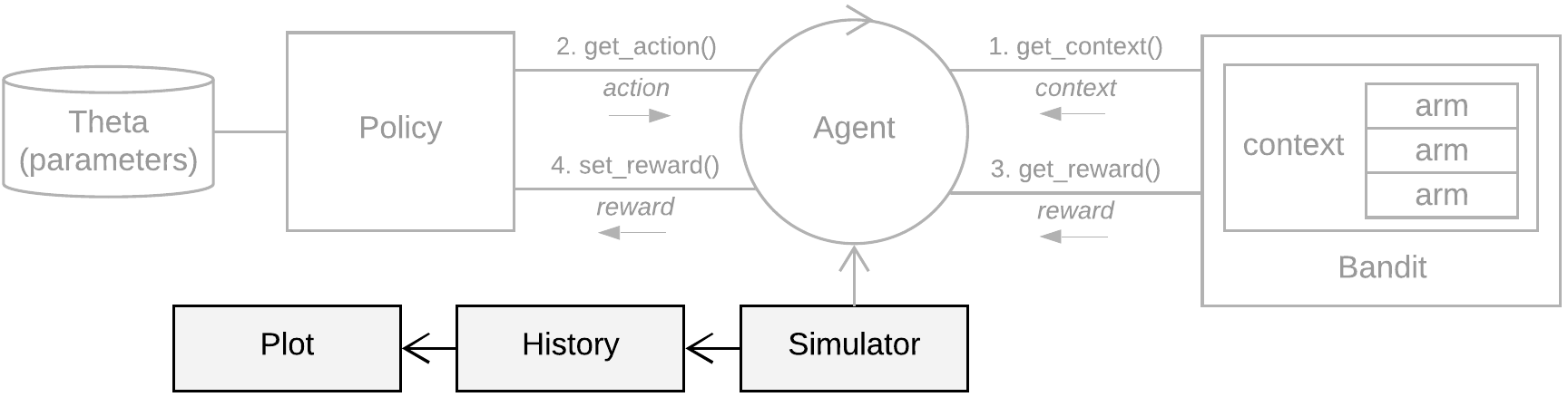}

The \code{Plot} class takes an \code{History} object and offers several ways to plot it, each optimized to be able to plot gigabytes worth of data, quickly:

\begin{itemize}
         \item \code{average}: plots the average reward or regret over all simulations per \code{Agent} instance (that is, over each \code{Bandit} and \code{Policy} instance 
	 combo) over time.
         \item \code{cumulative}: plots the average reward or regret over all simulations per \code{Agent} instance over time.
         \item \code{arms}: plots ratio of arms chosen on average at each time step, in percentages, totaling 100%.

\end{itemize}

\code{Plot} objects can be instantiated directly, or, more commonly, by calling the \code{plot()} function. In either case, make sure to specify a \code{History} instance 
and one of the plot types specified above:

\begin{Code}
# plot a history object through default generic plot() function
plot(history, type = "arms")

# or call the Plot() directly
p1 <- Plot$new()$cumulative(history)
p2 <- Plot$new()$average(history)
\end{Code}

Multiple \code{Agent} instances can be displayed within one \code{Plot}, and multiple plots can themselves again be combined into one graph\footnote{To do so, call 
\code{plot()} with \code{no\_par = TRUE}. This enables the setting of custom plotting parameters through \proglang{R}'s default \code{par()} functionality, allowing the 
formatting, layout and combination of single or multiple plots.}. Some example plots that illustrate many of \code{Plot()}'s features:

\begin{Code}
weights    <- matrix(c(0.7, 0.2, 0.2), 1, 3)

bandit     <- ContextualBernoulliBandit$new(weights = weights)

agents     <- list(Agent$new(RandomPolicy$new(), bandit),
                   Agent$new(OraclePolicy$new(), bandit),
                   Agent$new(ThompsonSamplingPolicy$new(1.0, 1.0), bandit),
                   Agent$new(Exp3Policy$new(0.1), bandit),
                   Agent$new(GittinsBrezziLaiPolicy$new(), bandit),
                   Agent$new(UCB1Policy$new(), bandit))

history    <- Simulator$new(agents, horizon = 100, simulations = 1000)$run()

plot(history, type = "cumulative", use_colors = FALSE, legend_border = FALSE,
     limit_agents = c("GittinsBrezziLai", "UCB1","ThompsonSampling"))

plot(history, type = "cumulative", regret = FALSE, legend = FALSE,
     limit_agents = c("Exp3"), traces = TRUE, no_par = TRUE)

plot(history, type = "cumulative", regret = FALSE, rate = TRUE,
     limit_agents = c("Exp3", "ThompsonSampling"), disp = "sd",
     legend_position = "bottomright", legend_border = FALSE)

plot(history, type = "cumulative", rate = TRUE, plot_only_disp = TRUE,
     disp = "var", smooth = TRUE, legend_position = "bottomleft",
     limit_agents = c("Exp3", "ThompsonSampling"), legend_border = FALSE)

plot(history, type = "average", disp = "ci", regret = FALSE, interval = 10,
     smooth = TRUE, legend_position = "bottomright", legend = FALSE)

plot(history, limit_agents = c("ThompsonSampling"), type = "arms",
     interval = 20)
\end{Code}
\begin{figure}[H]
\centering
\includegraphics[width=.99\textwidth]{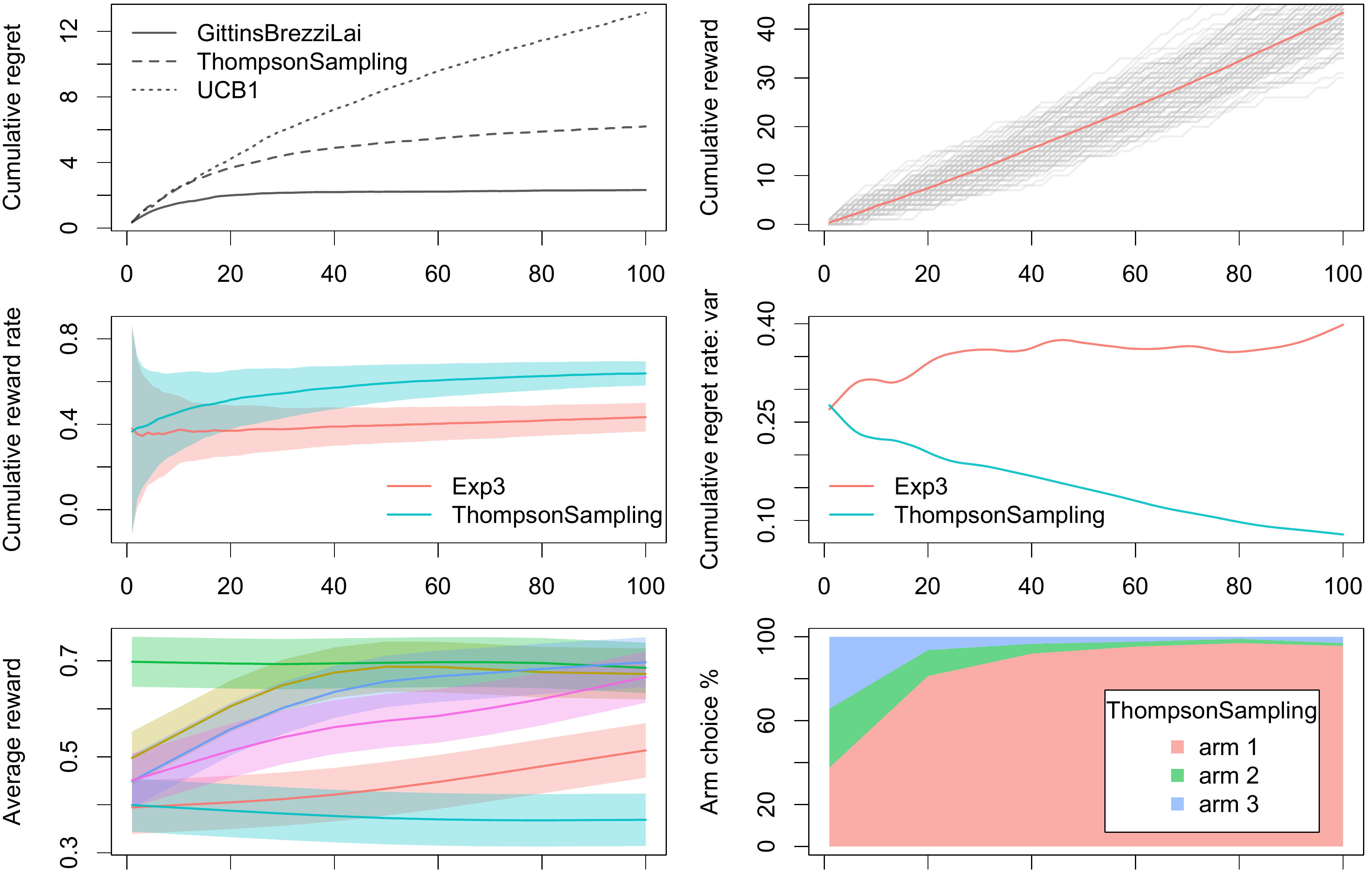}
\caption{Overview hightlighting the versatility of contextual's Plot() class. From left to right, and top to bottom: 1. Grayscale plot 2. Plotting individual simulation 
traces 3. Cumulative reward with standard deviation over simulations 4. Dispersion plot (here, variance) 5. Color plot with 95\% confidence interval 6. Arm choice percentage 
plot. }
\label{fig:section_4_2_plot}
\end{figure}

\section{Custom bandits and policies} \label{extending}

The current section illustrates how to develop custom \code{Bandit} and \code{Policy} subclasses through an exploration of \pkg{contextual}'s 
\code{ContextualBernoulliBandit} and its \code{EpsilonFirst}, \code{EspilonGreedy}, and \code{LinUCB} policy implementations.

\subsection{ContextualBernoulliBandit: a minimal contextual Bernoulli bandit} \label{ContextualBernoulliBandit}

Where not otherwise noted, all \code{Bandit} implementations in the current paper refer to (or will be configured as) multi-armed \code{Bandits} with Bernoulli rewards. For 
Bernoulli \code{Bandits}, the reward received is either a zero or a one: on each $t$ they offer either a reward of $1$ with probability $p$ or a reward of $0$ with 
probability $1 - p$. In other words, a Bernoulli bandit has a finite set of arms k \(  \in \left\{ 1, \dots, K \right\} \) where the rewards for each arm $k$ is distributed 
Bernoulli with parameter $\theta_k$ for the expected reward of the arm.

One example of a very simple contextual Bernoulli bandit is \pkg{contextual}'s basic weight-based contextual \code{Bandit} implementation, \code{ContextualBernoulliBandit}:

\begin{Code}
ContextualBernoulliBandit <- R6::R6Class(
  inherit = Bandit,
  class = FALSE,
  public = list(
    weights = NULL,
    class_name = "ContextualBernoulliBandit",
    initialize = function(weights) {
      self$weights     <- weights        # d x k weight matrix
      self$d           <- nrow(weights)  # d features
      self$k           <- ncol(weights)  # k arms
    },
    get_context = function(t) {
      # generate d dimensional feature vector, one random feature active at a time
      Xa <- sample(c(1,rep(0,self$d-1)))
      # convert to d x k matrix: one feature vector, recycled to every arm
      X  <- matrix(Xa, self$d, self$k)
      context <- list(
        X = X,
        k = self$k,
        d = self$d
      )
    },
    get_reward = function(t, context, action) {
      # which arm was selected?
      arm            <- action$choice
      # d dimensional feature vector for chosen arm
      Xa             <- context$X
      # weights of active context
      weight         <- Xa %*% self$weights
      # assign rewards for active context with weighted probs
      rewards        <- as.double(weight > runif(self$k))
      optimal_arm    <- which_max_tied(weight)
      reward  <- list(
        reward                   = rewards[arm],
        optimal_arm              = optimal_arm,
        optimal_reward           = weight[optimal_arm]
      )
    }
  )
)
\end{Code}

\code{ContextualBernoulliBandit} expects a  $d \times k$ \code{weights} matrix, where every element in \code{weights} represents the average probability of 
\code{ContextualBernoulliBandit} returning a reward of $1$ over each of its \code{k} arms as dependent on the currently active contextual feature.

\subsection{EpsilonFirstPolicy} \label{epsfirst}

An important feature of \pkg{contextual} is that it eases the conversion from formal and pseudocode policy descriptions to clean R6 classes. We will give several examples of 
such conversions in the current paper, starting with the implementation of the $\epsilon$-first algorithm. In this context-free, "naive" policy, also known as AB(C) testing, 
a pure exploration phase is followed by a pure exploitation phase.

In that respect, the $\epsilon$-first algorithm is equivalent to a randomized controlled trial (RCT). An RCT, generally referred to as the gold standard clinical research 
paradigm, is a study design where subjects are allocated at random to receive one of several clinical interventions. On completion of an RCT, the interventions are compared. 
If one intervention proves significantly better than the others, that intervention is suggested to be the superior "evidence-based" option from then on.

In other words: an $\epsilon$-first policy starts out by exploring arms at random during the first $\epsilon \cdot N$ time steps. The following $(1-\epsilon) \cdot N$ steps, 
$\epsilon$-first fully exploits the arm that proved the best during the previous $\epsilon \cdot N$ exploration phase. In pseudocode:

\begin{algorithm}[H]
\caption{$\epsilon$-first}
\label{Alg:EpsilonFirst}
\begin{algorithmic}
\REQUIRE \(   N \in \mathbb{Z}^{+} \) - horizon of the experiment.
\REQUIRE \(   \epsilon  \in \left[ 0,1 \right] \) - exploration tuning parameter
\STATE \( n_{k} \leftarrow 0 \) for all actions a \(  \in \left\{ 1, \dots, \mathcal{A} \right\} \)  (count how many times an arm has been chosen)
\STATE \( \hat{\mu}_{k} \leftarrow 0 \) for all actions a  \(   \in \left\{ 1, \dots, \mathcal{A} \right\} \)  (estimate of expected reward per arm)
% Run through time points:
\FOR{$t=1, \dots, T$}
	% Run through arms. Step 1, select which one to play
	\IF {\(t \leq \epsilon \cdot N\)}
	       \STATE play a random arm out of all actions a \(   \in \left\{ 1, \dots, \mathcal{A} \right\} \)
	\ELSE
	        \STATE play action \(a_t = \argmax_a  \hat{\mu}_{t=\eta,a}  \) with ties broken arbitrarily
	\ENDIF
	\STATE observe real-valued payoff $r_t$
	% Update:
	\STATE \( n_{a_{t}} \leftarrow n_{a_{t-1}} + 1  \) \COMMENT{Update count}
  \STATE \( \hat{\mu}_{t,a_{t}} \leftarrow   \cfrac{r_t - \hat{\mu}_{t-1,a_{t}} }{n_{a_{t}}}   \) \COMMENT{Update expected reward}
\ENDFOR
\end{algorithmic}
\end{algorithm}

And the above pseudocode converted to an \code{EpsilonFirstPolicy} class:

\begin{Code}
EpsilonFirstPolicy              <- R6::R6Class(
  portable = FALSE,
  class = FALSE,
  inherit = Policy,
  public = list(
    first = NULL,
    class_name = "EpsilonFirstPolicy",
    initialize = function(epsilon = 0.1, N = 1000, time_steps = NULL) {
      super$initialize()
      self$first                <- ceiling(epsilon*N)
      if (!is.null(time_steps)) self$first <- time_steps
    },
    set_parameters = function(context_params) {
      self$theta_to_arms        <- list('n' = 0, 'mean' = 0)
      # Here we define a list with 'n' and 'mean' theta parameters to each
      # arm through helper variable self$theta_to_arms. That is, when the
      # number of arms is 'k', the above would equal:

      # self$theta <- list(n = rep(list(0,k)), 'mean' = rep(list(0,k)))

      # ... which would also work just fine, but is much less concise.

      # When assigning both to self$theta directly & via self$theta_to_arms,
      # make sure to do it in that particular order.
    },
    get_action = function(t, context) {
      if (sum_of(self$theta$n) < self$first) {
        action$choice           <- sample.int(context$k, 1, replace = TRUE)
        action$propensity       <- (1/context$k)
      } else {
        action$choice           <- which_max_list(self$theta$mean)
        action$propensity       <- 1
      }
      action
    },
    set_reward = function(t, context, action, reward) {
      arm                       <- action$choice
      reward                    <- reward$reward
      inc(self$theta$n[[arm]])  <- 1
      if (sum_of(self$theta$n) < self$first - 1) {
        inc(self$theta$mean[[arm]]) <- 
            (reward - self$theta$mean[[arm]]) / self$theta$n[[arm]]
      }
      self$theta
    }
  )
)
\end{Code}

To evaluate this policy, instantiate both an \code{EpsilonFirstPolicy} and a \code{ContextualBernoulliBandit}. Then add the \code{Bandit}/\code{Policy} pair to an 
\code{Agent}. Next, add the \code{Agent} to a \code{Simulator}. Finally, run the \code{Simulator}, and \code{plot()} the its \code{History} log:

\begin{Code}
horizon            <- 100
simulations        <- 1000
weights            <- matrix(c(0.6, 0.2, 0.2), 1, 3)

policy             <- EpsilonFirstPolicy$new(epsilon = 0.5, N = horizon)
bandit             <- ContextualBernoulliBandit$new(weights = weights)

agent              <- Agent$new(policy,bandit)

simulator          <- Simulator$new(agents = agent,
                                    horizon = horizon,
                                    simulations = simulations)

history            <- simulator$run()

plot(history, type = "cumulative")
plot(history, type = "arms")
\end{Code}
\begin{figure}[H]
\centering
\includegraphics[width=.99\textwidth]{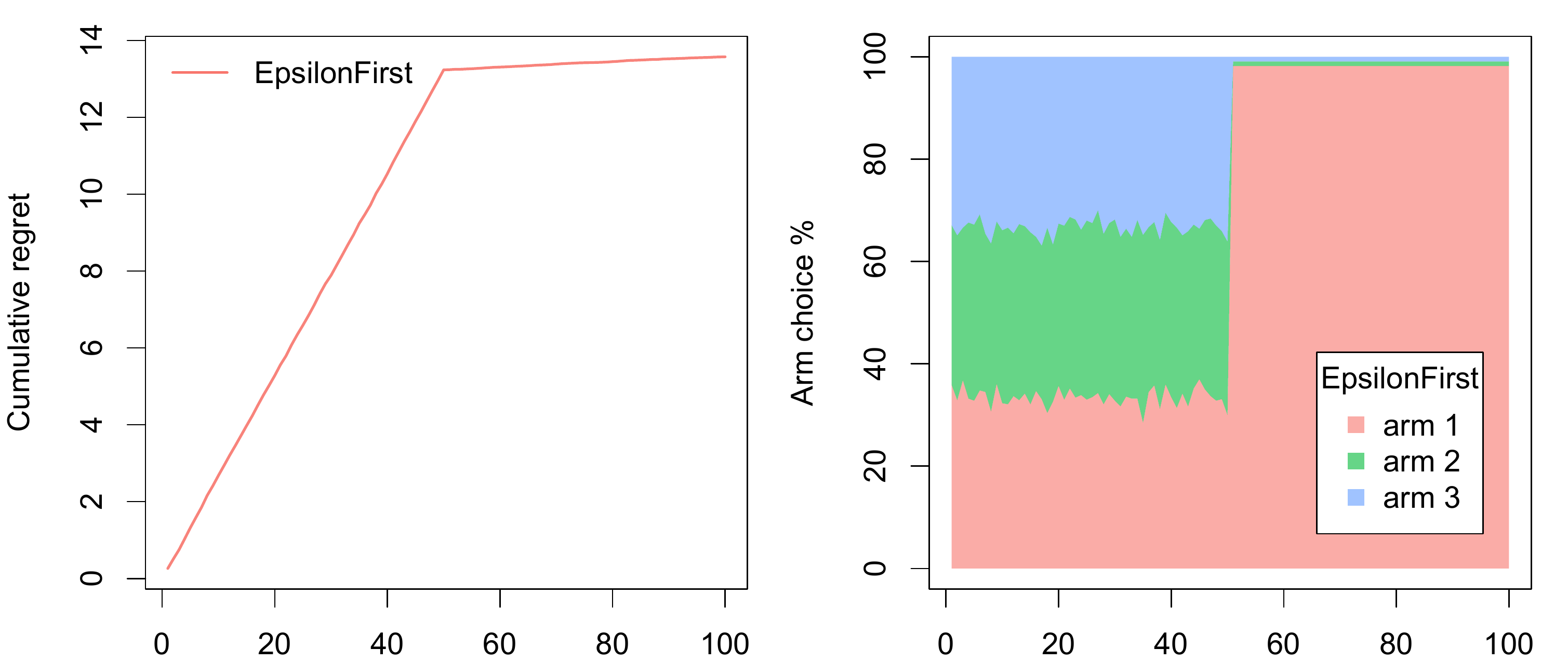}
\caption{To the left, Epsilon First policy's cumulative regret over time. To the right, the percentage of simulations per time step for which each of the bandit's three arms 
where chosen. As can be observed in the arm choice plot, after fifty steps of exploration, the policy has settled overwhelmingly on arm one. On exploiting this arm, from 
thereon, it suffers significantly less regret than before.}
\label{fig:section_5_2}
\end{figure}

\subsection{EpsilonGreedyPolicy} \label{epsgreedy}

Contrary to the previously introduced $\epsilon$-first policy, an $\epsilon$-greedy algorithm \citep{Sutton1998e} does not divide exploitation and exploration into two 
strictly separate phases---it explores with a probability of $\epsilon$ and exploits with a probability of $1-\epsilon$, right from the start. That is, an $\epsilon$-greedy 
policy with an $\epsilon$ of $0.1$ explores arms at random 10\% of the time. The other $1-\epsilon$, or 90\% of the time, the policy "greedily" exploits the currently 
best-known arm.

This can be formalized in pseudocode as follows:

\begin{algorithm}[H]
\caption{$\epsilon$-greedy}
\label{Alg:EpsilonGreedy}
\begin{algorithmic}
\REQUIRE \(    \epsilon  \in \left[ 0,1 \right] \) - exploration tuning parameter
\STATE \( n_{a} \leftarrow 0 \) for all actions a \(  \in \left\{ 1, \dots, \mathcal{A} \right\} \)  (count how many times an action has been chosen)
\STATE \( \hat{\mu}_{a} \leftarrow 0 \) for all actions a  \(   \in \left\{ 1, \dots, \mathcal{A} \right\} \)  (estimate of expected reward per action)
% Run through time points:
\FOR{$t=1, \dots, T$}
	% Run through actions. Step 1, select which one to play
	\IF {sample from $unif(0,1) > \epsilon$}
		\STATE play action \(a_t = \argmax_a  \hat{\mu}_{t-1,a}  \) with ties broken arbitrarily
	\ELSE
		\STATE play a random action out of all actions a \(  \in \left\{ 1, \dots, \mathcal{A} \right\} \)
	\ENDIF
	\STATE observe real-valued payoff $r_t$
	% Update:
	\STATE \( n_{a_{t}} \leftarrow n_{a_{t-1}} + 1  \) \COMMENT{Update count}
  \STATE \( \hat{\mu}_{t,a_{t}} \leftarrow   \cfrac{r_t - \hat{\mu}_{t-1,a_{t}} }{n_{a_{t}}}   \) \COMMENT{Update expected reward}
\ENDFOR
\end{algorithmic}
\end{algorithm}

Converted to an EpsilonGreedyPolicy class:

\begin{Code}
#' @export
EpsilonGreedyPolicy          <- R6::R6Class(
  portable = FALSE,
  class = FALSE,
  inherit = Policy,
  public = list(
    epsilon = NULL,
    class_name = "EpsilonGreedyPolicy",
    initialize = function(epsilon = 0.1) {
      super$initialize()
      self$epsilon                <- epsilon
    },
    set_parameters = function(context_params) {
      self$theta                  <- list('exploit' = 0)
      self$theta_to_arms          <- list('n' = 0, 'mean' = 0)
    },
    get_action = function(t, context) {
      if (runif(1) > self$epsilon) {
        # exploit best arm
        self$theta$exploit        <- 1
        self$action$choice        <- which_max_list(self$theta$mean)
        self$action$propensity    <- 1 - self$epsilon
      } else {
        # explore any arm
        self$theta$exploit        <- 0
        self$action$choice        <- sample.int(context$k, 1, replace = TRUE)
        self$action$propensity    <- self$epsilon*(1/context$k)
      }
      self$action
    },
    set_reward = function(t, context, action, reward) {
      arm                         <- action$choice
      reward                      <- reward$reward
      self$theta$n[[arm]]         <- self$theta$n[[arm]] + 1
      self$theta$mean[[arm]]      <- self$theta$mean[[arm]] + 
          (reward - self$theta$mean[[arm]]) / self$theta$n[[arm]]
      self$theta
    }
  )
)
\end{Code}
\begin{figure}[H]
\centering
\includegraphics[width=.99\textwidth]{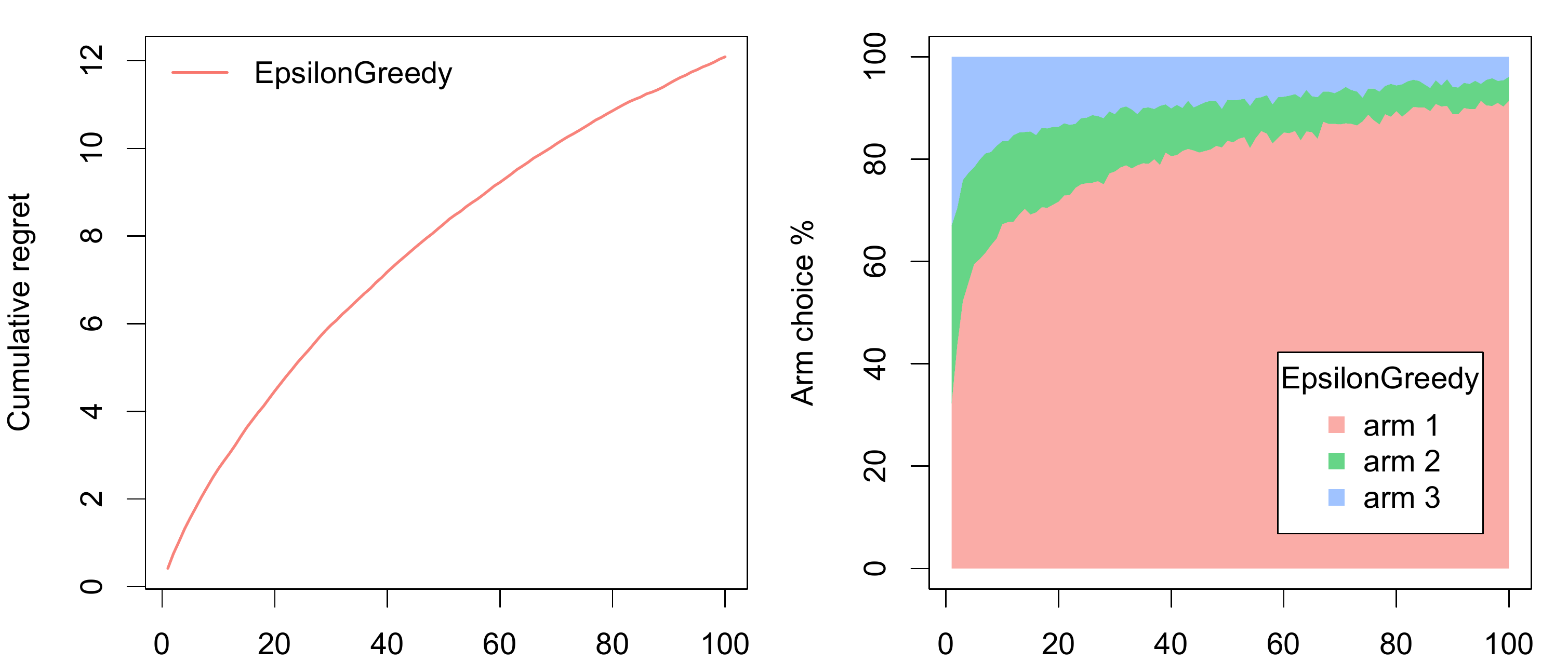}
\caption{To the left, Epsilon Greedy policy's cumulative regret over time. To the right, the percentage of simulations per time step for which each of the bandit's three 
arms where chosen. The Epsilon Greedy policy starts to both explore and exploit right from the start, gradually choosing the best arm more and more.}
\label{fig:section_5_3}
\end{figure}

Assign the new class, together with \code{ContextualBernoulliBandit}, to an \code{Agent}. Again, assign the \code{Agent} to a \code{Simulator}. Then run the \code{Simulator} 
and \code{plot()}:

\begin{Code}
horizon            <- 100
simulations        <- 1000
weights            <- matrix(c(0.8, 0.2, 0.2), 1, 3)

policy             <- EpsilonGreedyPolicy$new(epsilon = 0.1)
bandit             <- ContextualBernoulliBandit$new(weights = weights)

agent              <- Agent$new(policy,bandit)

simulator          <- Simulator$new(agents = agent,
                                    horizon = horizon,
                                    simulations = simulations)

history            <- simulator$run()

plot(history, type = "cumulative")
plot(history, type = "arms")
\end{Code}

\subsection{Contextual Bandit: LinUCB with linear disjoint models} \label{linucbc}

As a final example of how to subclass \pkg{contextual}'s \code{Bandit} superclass, we move from context-free algorithms to a contextual one. As described in Section 
\ref{intro}, contextual bandits make use of side information to help them choose the current best arm to play. For example, contextual information such as a website 
visitors' location may be related to which article's headline (or arm) on the frontpage of the website will be clicked on most.

Here, we show how to implement and evaluate probably one of the most cited out of all contextual policies, the LinUCB algorithm with Linear Disjoint Models \cite{Li2010}. 
The policy is more complicated than the previous two bandits, but when following its pseudocode description to the letter, it translates nicely to yet another \code{Bandit} 
subclass.

The LinUCB algorithm employs ridge regression with feature vectors and rewards per arm to predict rewards using the trials performed so far. Importantly, the algorithm also 
generates a confidence interval for this predicted payoff. The policy then chooses the arm with the highest upper confidence bound. In pseudocode, following Algorithm 1 from 
\cite{Li2010}:

\begin{algorithm}[H]
\caption{LinUCB with linear disjoint models}
\label{Alg:LinUCBDisjoint}
\begin{algorithmic}
\REQUIRE $\alpha$ \(  \in \mathbb{R}^{+} \), exploration tuning parameter
% Run through time points:
\FOR{$t=1, \dots, T$}
          \STATE Observe features of all actions \(  a \in \mathcal{A}_{t}: x_{t,a} \in \mathbb{R}^{d}\)
	% Run through actions. Step 1, select which one to play
	\FOR{ \(  a \in \mathcal{A}\)}
	          \IF{\(a\) is new}
		      \STATE \(A_{a} \leftarrow I_{d}  \)  (d-dimensional identity matrix)
		      \STATE \(b_{a} \leftarrow 0_{d\times1}   \) (d-dimensional zero vector)
		\ENDIF
		\STATE \( \hat{\theta}_{a} \leftarrow A_{a}^{-1}b_{a} \)
		\STATE \( p_{t,a} \leftarrow \hat{\theta}_{a}^{T} + \alpha  \sqrt{ x_{t,a}^{T} A_{a}^{-1}x_{t,a}} \)
	\ENDFOR
	% allocate to action
	\STATE Play action \(a_t = \argmax_a  p_{t,a}  \) with ties broken arbitrarily and observe real-valued payoff $r_t$
	% Update:
           \STATE \( A_{a_{t}} \leftarrow A_{a_{t}}+ x_{t,a_{t}}x_{t,a_{t}}^{T} \)
           \STATE  \( b_{a_{t}} \leftarrow b_{a_{t}}+ r_{t}x_{t,a_{t}}  \)
\ENDFOR
\end{algorithmic}
\end{algorithm}

Next, translating the above pseudocode into a well organized \code{Bandit} subclass:

\begin{Code}
LinUCBDisjointPolicy <- R6::R6Class(
  portable = FALSE,
  class = FALSE,
  inherit = Policy,
  public = list(
    alpha = NULL,
    class_name = "LinUCBDisjointPolicy",
    initialize = function(alpha = 1.0) {
      super$initialize()
      self$alpha <- alpha
    },
    set_parameters = function(context_params) {
      ul <- length(context_params$unique)
      self$theta_to_arms <- list('A' = diag(1,ul,ul), 'b' = rep(0,ul))
    },
    get_action = function(t, context) {
      expected_rewards <- rep(0.0, context$k)
      for (arm in 1:context$k) {
        Xa         <- get_arm_context(context, arm, context$unique)
        A          <- self$theta$A[[arm]]
        b          <- self$theta$b[[arm]]
        A_inv      <- inv(A)
        theta_hat  <- A_inv %*% b
        mu_hat     <- Xa %*% theta_hat
        sigma_hat  <- sqrt(tcrossprod(Xa %*% A_inv, Xa))
        expected_rewards[arm] <- mu_hat + self$alpha * sigma_hat
      }
      action$choice  <- which_max_tied(expected_rewards)
      action
    },
    set_reward = function(t, context, action, reward) {
      arm    <- action$choice
      reward <- reward$reward
      Xa     <- get_arm_context(context, arm, context$unique)
      inc(self$theta$A[[arm]]) <- outer(Xa, Xa)
      inc(self$theta$b[[arm]]) <- reward * Xa
      self$theta
    }
  )
)
\end{Code}

\begin{figure}[H]
\centering
\includegraphics[width=1.0\textwidth]{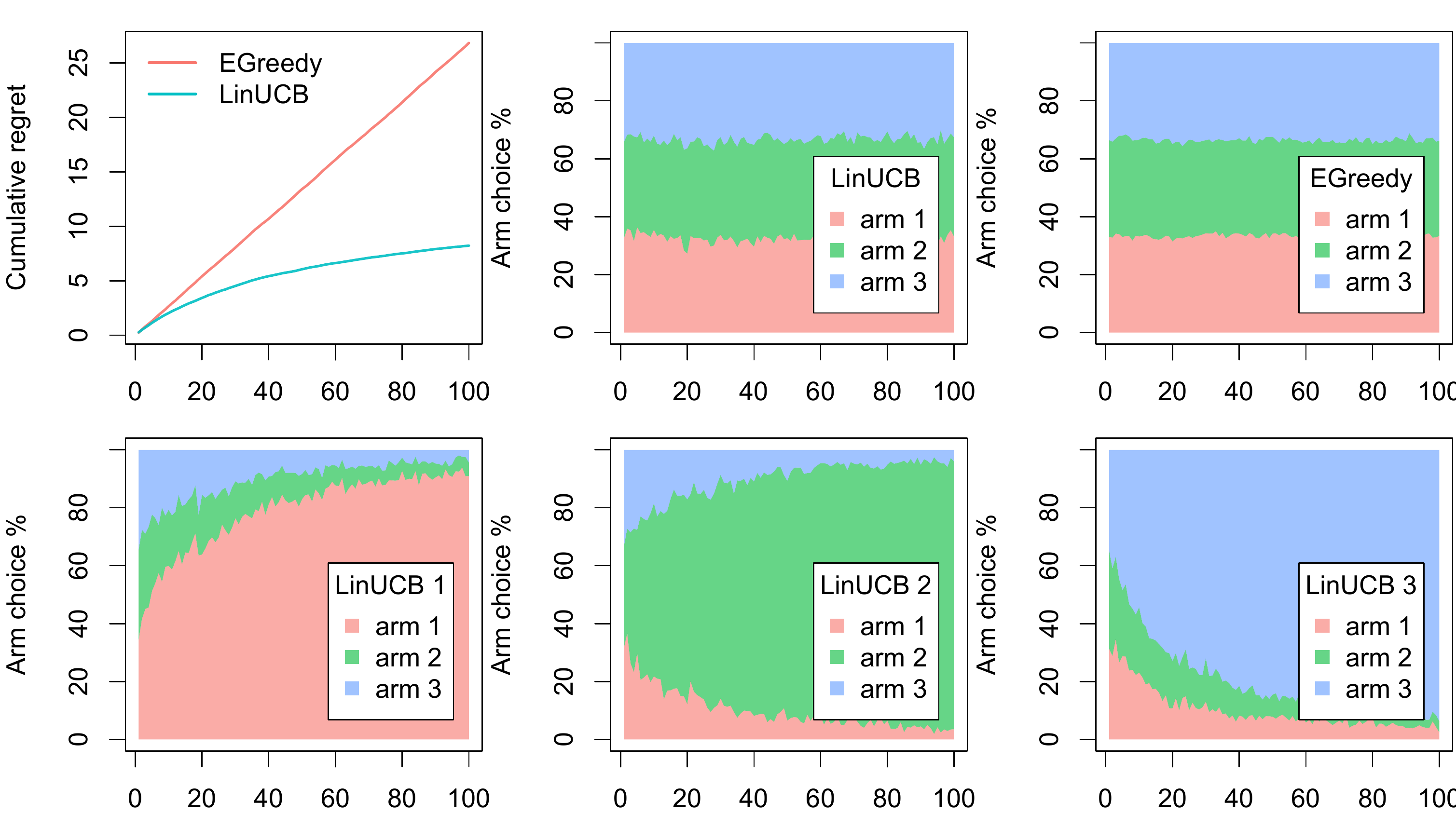}
\caption{Top left a plot of Epsilon Greedy versus LinUCB policies' cumulative regret over time. Top middle and right, Epsilon Greedy and LinUCB policies' average arm choice 
percentage per $t$, which indicate that, overall, both policies choose each arm about equally often. To see why LinUCB is able to do better than Epsilon Greedy, we need to 
look at the bottom row, whose three plots reperesent LinUCB's average arm choice percentage by context feature. These bottom row plots show how LinUCB is able to make use of 
contextual features to map actions to rewards.}
\label{fig:section_5_4}
\end{figure}

Let us now evaluate the above \code{LinUCBDisjointPolicy} using a Bernoulli \code{ContextualBernoulliBandit} with three arms and three context features. In the code below we 
define each of \code{ContextualBernoulliBandit}'s arms to be, on average, equally probable to return a reward. As can be confirmed in the two plots to the right below,  
where \code{EpsilonGreedyPolicy} and \code{LinUCBDisjointPolicy} actually show no difference in their overall arm preference. Yet \code{LinUCBDisjointPolicy}'s awareness of 
the changing context allows it to learn the relationships between arms, rewards, and features, and select the right arm in the right context---as can be observed in the 
bottom row of Figure \ref{fig:section_5_4}.

\begin{Code}

horizon     <- 100L
simulations <- 1000L

                      # k=1  k=2  k=3             -> columns represent arms
weights     <- matrix(c(0.6, 0.2, 0.2,     # d=1  -> rows represent
                        0.2, 0.6, 0.2,     # d=2     context features,
                        0.2, 0.2, 0.6),    # d=3

                      nrow = 3, ncol = 3, byrow = TRUE)

bandit      <- ContextualBernoulliBandit$new(weights = weights)

eg_policy   <- EpsilonGreedyPolicy$new(0.1)
lucb_policy <- LinUCBDisjointPolicy$new(0.6)

agents      <- list(Agent$new(eg_policy, bandit, "EGreedy"),
                    Agent$new(lucb_policy, bandit, "LinUCB"))

simulation  <- Simulator$new(agents, horizon, simulations, save_context = TRUE)
history     <- simulation$run()

plot(history, type="cumulative", legend_border = FALSE)
plot(history, type="arms", limit_agents = c("LinUCB"))
plot(history, type="arms", limit_agents = c("EGreedy"))

plot(history, type="arms", limit_agents = c("LinUCB"), limit_context = c("X.1"))
plot(history, type="arms", limit_agents = c("LinUCB"), limit_context = c("X.2"))
plot(history, type="arms", limit_agents = c("LinUCB"), limit_context = c("X.3"))

\end{Code}

\section{Subclassing bandits and policies} \label{subclpb}

The \pkg{contextual} package's extensibility does not limit itself to the subclassing of \code{Policy} classes. Through its R6 based object system it is easy to extend and 
override any \pkg{contextual} super- or subclass. Below, we demonstrate how to apply that extensibility to sub-subclass one \code{Bandit} and one \code{Policy} subclass.
First, we extend \pkg{contextual}'s context-free \code{BasicBernoulliBandit}, replacing its Bernoulli based reward function with a Poisson based one \citep{Presman1991}. 
Next, we implement an \code{EpsilonGreedyAnnealingPolicy} version of the $\epsilon$-greedy policy introduced in Section \ref{epsgreedy}---where its 
\code{EpsilonGreedyAnnealingPolicy} subclass introduces a gradual reduction ("annealing") of the policy's $epsilon$ parameter over T 
\citep{Cesa-Bianchi1998,Kirkpatrick1983}, in effect making the policy more exploitative over time.

\begin{Code}
BasicPoissonBandit <- R6::R6Class(
  inherit = BasicBernoulliBandit,
  class = FALSE,
  public = list(
    weights = NULL,
    class_name = "BasicPoissonBandit",
    # Override get_reward & generate Poisson based rewards
    get_reward = function(t, context, action) {
      reward_means = rep(2,self$k)
      rpm <- rpois(self$k, reward_means)
      rewards <- matrix(rpm < self$weights, self$k, 1)*1
      optimal_arm    <- which_max_tied(self$weights)
      reward         <- list(
        reward                   = rewards[action$choice],
        optimal_arm              = optimal_arm,
        optimal_reward           = rewards[optimal_arm]
      )
    }
  )
)

EpsilonGreedyAnnealingPolicy <- R6::R6Class(
  # Class extends EpsilonGreedyPolicy
  inherit = EpsilonGreedyPolicy,
  portable = FALSE,
  public = list(
    class_name = "EpsilonGreedyAnnealingPolicy",
    # Override EpsilonGreedyPolicy's get_action, use annealing epsilon
    get_action = function(t, context) {
      self$epsilon <- 1/(log(100*t+0.001))
      super$get_action(t, context)
    }
  )
)

weights <- c(7,1,2)
horizon <- 200
simulations <- 1000
bandit <- BasicPoissonBandit$new(weights)
ega_policy <- EpsilonGreedyAnnealingPolicy$new()
eg_policy  <- EpsilonGreedyPolicy$new(0.2)
agents <- list(Agent$new(ega_policy, bandit, "EG Annealing"),
               Agent$new(eg_policy, bandit, "EG"))
simulation <- Simulator$new(agents, horizon, simulations, do_parallel = FALSE)
history <- simulation$run()


par(mfrow = c(1, 3), mar = c(2, 4, 1, 0.1), cex=1.3)  #bottom, left, top, and right.

plot(history, type = "cumulative", no_par = TRUE, legend_border = FALSE,
              legend_position = "bottomright")
plot(history, type = "arms",  limit_agents = c("EG Annealing"), no_par = TRUE,
              interval = 25)
plot(history, type = "arms",  limit_agents = c("EG"), no_par = TRUE,
              interval = 25)
\end{Code}
\begin{figure}[H]
\centering
\includegraphics[width=.99\textwidth]{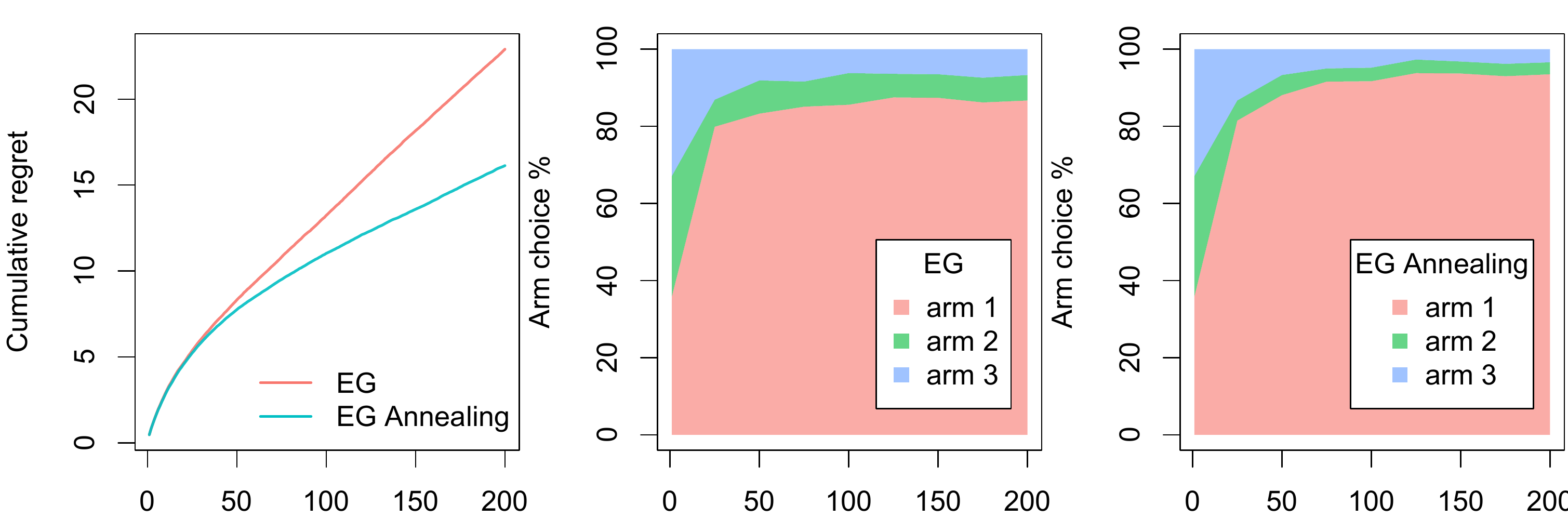}
\caption{To the left, both policies' cumulative regret over time. Middle and right, the percentage of simulations per time step for which each of the bandit's three arms 
where chosen. In the rightmost plot, it can be observed that, in contrast to its non-annealing cousin, the annealing policy is able to explore the sub-optimal arms less and 
less over time.}
\label{fig:section_5_2}
\end{figure}

\section{Offline evaluation} \label{offl}

As demonstrated in the previous section, it is relatively easy to evaluate MAB and CMAB policies against basic synthetic bandits. Still, the creation of more elaborate 
synthetic bandits can become very difficult very fast \citep{Strehl2006a}. Moreover, even when these technical hurdles are surmounted, synthetic bandits generally remain 
biased with respect to the real world interactions they aim to model \citep{Li2012,Li2011}.

One way to overcome these limitations would be to evaluate policies in a live setting. Online evaluations can deliver unbiased, realistic estimates of a policy's 
effectiveness. They are, however, generally much slower than synthetic evaluation, as online evaluations are dependent on active user interventions \citep{Mandel2016, 
Tewari2017}. Also, testing of potentially suboptimal policies on live audiences, such as patients or customers, can be dangerous, expensive, or even unethical 
\citep{Bastani2015}.

Fortunately, there is a third option: the evaluation of policies on logged bandit feedback. Such logs are widely available to little or no cost\footnote{Generally, 
interaction logs already generated by default in many types of interactive systems, from online search to news recommendation and online advertising.}, and their use in the 
"offline" or "batched" evaluation of policies does not have any of the potential adverse effects of online testing. There is a catch, however: With logged interaction data, 
we only have access to rewards awarded to actions that were chosen by some other policy at a previous time---but we are interested in the rewards for choices made by the 
policy under evaluation right now. In other words, such logs only contain partial feedback with respect to the policy under evaluation \citep{Strehl2010}.

So we need some bandit that can evaluate policies on interaction logs that were generated by another policy at another time. Such a bandit would need to be:

\begin{itemize}
   \item{Off-policy: The bandit has to be able to compare the policy under evaluation against data generated by another policy at another time \citep{Li2012,Li2011}.
   }
   \item{Counterfactual: The bandit needs some strategy for when the policy under evaluation chooses a different, "counterfactual" action from the one chosen by the logged 
   policy \citep{Bottou2013,Swaminathan2015}.}
\end{itemize}

In the next section, we introduce \pkg{contextual}'s implementation of \cite{Li2011}'s offline, off-policy "replay" \citep{Nicol2014} evaluator as a first example of such a 
"counterfactual" bandit.

\subsection{The replay method} \label{offli}

One approach to the use of offline data in off-policy evaluation is to recognize that we need to limit our evaluation to those rows of data where the action taken by the 
original logging policy equals the action suggested by the policy under evaluation \citep{Li2012,Li2011}. Such a so-called "replay" type bandit goes over every logged row 
\((x_{t},a_{t},r_{t})\), and only uses those (on average) 1/K rows where the action chosen by the policy under evaluation is the same as the original, off-policy action 
\citep{Nicol2014}.

In other words, for each $t$ in $T$, a replay bandit:

\begin{itemize}
   \item{Retrieves initially logged context $x_{t}$ from $D$ and makes it available to the policy under evaluation.}
   \item{Tests whether the action $a_{t}$ as chosen by the policy under evaluation equals the one that was logged in $D$.}
   \item{If both actions are the same, the bandit makes the originally logged reward available to the policy under evaluation. If not, the current row of $D$ is not taken 
   into account. }
\end{itemize}

In pseudocode, following Algorithm 2 from \cite{Li2011}:

\begin{algorithm}[H]
\caption{Replay Policy Evaluator}
\label{Alg:LiBandit}
\begin{algorithmic}
\REQUIRE  Policy $\piup$ \\
                 Data stream of randomly allocated events $S$ of length $T$  \\
                 $h_0 \leftarrow \emptyset$ {An initially empty history log}\\
                 $R_\pi \leftarrow 0$ {An initially zero total cumulative reward}\\
                 $L \leftarrow 0$ {An initially zero length counter of valid events}
% Run through time points:
\FOR{$t=1, \dots, T$}
	\STATE Get the $t$-th event \( (x_{t,a_t},a_{t},r_{t,a_t}) \) from  $S$
	\IF {\(\pi \left( h_{t-1},x_{t,a_t} \right) = a_t\)}
	       \STATE $h_{t} \leftarrow $  \(\textrm{CONCATENATE}\left( h_{t-1},(x_{t,a_t},a_{t},r_{t,a_t})  \right)\)
	       \STATE $R_\pi = R_\pi + r_{t,a_t}$
	       \STATE $L = L + 1$
	\ELSE
	        \STATE $h_{t} \leftarrow  h_{t-1} $
	\ENDIF
\ENDFOR
\STATE Output: rate of cumulative regret $R_\pi / L $
\end{algorithmic}
\end{algorithm}

Below, an implementation of this algorithm (a version of \pkg{contextual}'s  OfflineReplayEvaluatorBandit), followed by an offline evaluation and comparison of 
\pkg{contextual}'s \code{LinUCBDisjointPolicy} for four different values of its $\alpha$ parameter. We feed the replay bandit a personalization dataset\footnote{Made 
available by T. Lebera's for use in his "Machine Learning for Personalization" course at http://www.cs.columbia.edu/~jebara/6998/} consisting of a text file with 10000 rows, 
each row consisting of 102 space separated columns. The first of these columns represents which out of ten actions was performed. The second of the columns is a binary 
reward, and the remaining columns constitute the $x_t \in \mathbb{R}^{100}$ context vector, stored as integers for numerical efficiency \citep{lebera}.

\begin{Code}
library(contextual)
library(data.table)

# Define Replay Bandit
OfflineReplayEvaluatorBandit <- R6::R6Class(
  inherit = Bandit,
  private = list(
    S = NULL
  ),
  public = list(
    class_name = "OfflineReplayEvaluatorBandit",
    initialize   = function(offline_data, k, d) {
      self$k <- k                 # Number of arms
      self$d <- d                 # Context feature vector dimensions
      private$S <- offline_data   # Logged events
    },
    get_context = function(index) {
      context <- list(
        k = self$k,
        d = self$d,
        X = private$S$context[[index]]
      )
      context
    },
    get_reward = function(index, context, action) {
      if (private$S$choice[[index]] == action$choice) {
        list(
          reward = as.double(private$S$reward[[index]])
        )
      } else {
        NULL
      }
    }
  )
)

# Import personalization data-set
url      <- "http://d1ie9wlkzugsxr.cloudfront.net/data_cmab_basic/dataset.txt"
datafile <- fread(url)

# Clean up datafile
datafile[, context := as.list(as.data.frame(t(datafile[, 3:102])))]
datafile[, (3:102) := NULL]
datafile[, t := .I]
datafile[, sim := 1]
datafile[, agent := "linucb"]
setnames(datafile, c("V1", "V2"), c("choice", "reward"))

# Set simulation parameters.
simulations <- 1
horizon     <- nrow(datafile)

# Initiate Replay bandit with 10 arms and 100 context dimensions
log_S       <- datafile
bandit      <- OfflineReplayEvaluatorBandit$new(log_S, k = 10, d = 100)

# Define agents.
agents      <-
  list(Agent$new(LinUCBDisjointOptimizedPolicy$new(0.01),bandit,"alpha = 0.01"),
       Agent$new(LinUCBDisjointOptimizedPolicy$new(0.05),bandit,"alpha = 0.05"),
       Agent$new(LinUCBDisjointOptimizedPolicy$new(0.1),bandit,"alpha = 0.1"),
       Agent$new(LinUCBDisjointOptimizedPolicy$new(1.0),bandit,"alpha = 1.0"))

# Initialize the simulation.
simulation  <-
  Simulator$new(
    agents           = agents,
    simulations      = simulations,
    horizon          = horizon,
    save_context     = TRUE
  )

# Run the simulation.
linucb_sim  <- simulation$run()

# Plot the results.
# plot the results
plot(linucb_sim, type = "cumulative", legend_title = "LinUCB",
     rate = TRUE, regret = FALSE, legend_position = "bottomright")
\end{Code}

\begin{figure}[H]
  \centering
    \includegraphics[width=.99\textwidth]{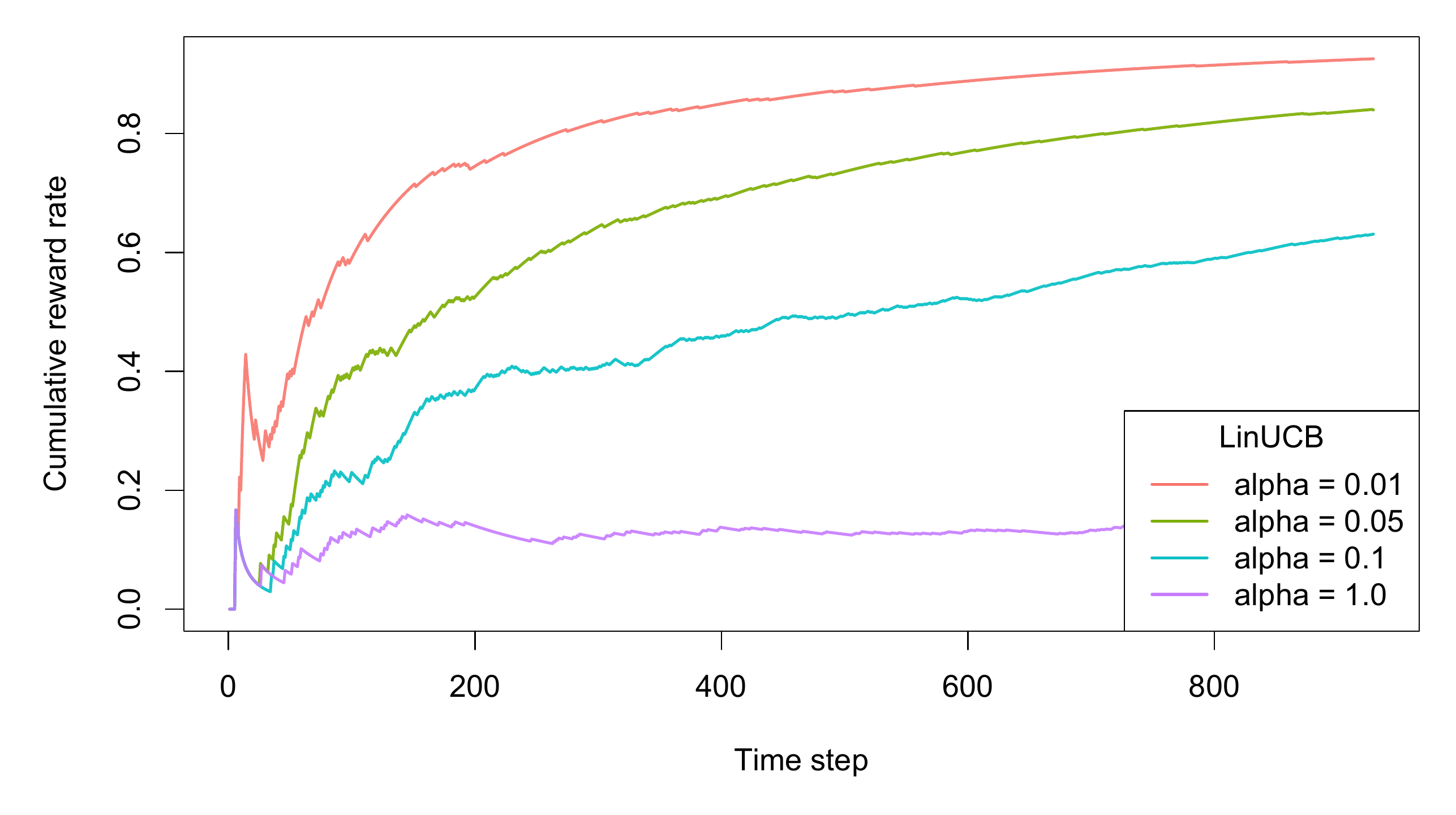}

      \caption{LinUCBHybridPolicy evaluated with OfflineReplayEvaluatorBandit. As the original logging policy randomly allocated actions to each of ten available arms, the 
      logged and evaluated policy's actions correspond to each other about $1/K = 1/10$ or 1,000 out of the original 10,0000 events.}
      \label{fig:offline_bandit}
\end{figure}

See Figure \ref{fig:offline_bandit} for a plot of the resulting cumulative reward rate over time.

The replay method works well for logs where the original logging policy randomly allocated actions over the available arms, $p(a) = 1/K$. Where this is not the case, 
however, data generating policies can additionally calculate probabilities $p_{t,a_t}$ of selecting an action, returning it as \code{action$propensity}, in addition to 
\code{action$choice} in its \code{get_action()} call. For instance, in \pkg{contextual}'s $\epsilon$-greedy implementation:

\begin{Code}
EpsilonGreedyPolicy          <- R6::R6Class(
    ...
    get_action = function(t, context) {
      if (runif(1) > self$epsilon) {
        self$action$choice        <- which_max_list(self$theta$mean)
        self$action$propensity    <- 1 - self$epsilon
      } else {
        self$action$choice        <- sample.int(context$k, 1, replace = TRUE)
        self$action$propensity    <- self$epsilon*(1/context$k)
      }
      self$action
    },
    ...
)
\end{Code}

This propensity can then:

\begin{itemize}
 \item{Be used for inverse propensity matching or weighting \cite{Austin2011} to estimate the action's causal effect by accounting for contextual covariates 
 \citep{Imbens2015,Pearl2009}. }

 \item{Stored in the log, resulting in \( D = (x_{t,a_t},a_{t},r_{t,a_t},p_{t,a_t}) \), and then be used to estimate average rewards using inverse propensity scoring 
 \citep{Horvitz1952} by computing:
\begin{eqnarray}
\texttt{ips}(\pi) & = & \frac{1}{N} \sum_{t=1}^{N} \mathbbm{1} \{ \pi(x_t) = a_t \} r_t / p_t
\end{eqnarray}

where the indicator is $1$ when $\pi$'s action matches the action in the logs \citep{kruijswijk2018streamingbandit}. See \pkg{contextual}'s 
\code{OfflinePropensityWeightingBandit} for a basic implementation of this mechanism.}
\end{itemize}

The current section is but a short introduction to offline policy evalution. For more information on inverse propensity scoring, doubly robust evaluation, and other batch or 
offline evaluation methods, see for example \citet{Agarwal2014}, \citet{Dudik2011} and \citet{Swaminathan2015}.

\section{Replication of Li et al 2010} \label{repl}

In the current section, we demonstrate how \pkg{contextual} facilitates the comparison of bandit policies on big offline datasets by running a partial replication of 
\cite{Li2010}. This paper describes how the authors made use of offline Yahoo! click-through rate data to evaluate and compare the effectiveness of several context-free and 
contextual policies---therein introducing both the offline policy evaluator outlined in the previous section and the LinUCB algorithm introduced in Section \ref{linucbc}.

\subsection{Description of the data} \label{datadesc}

The dataset used in the \cite{Li2010} paper has been made available at the Yahoo! lab's website\footnote{At 
https://webscope.sandbox.yahoo.com/catalog.php?datatype=r\&did=49}. It contains the click-through rate from the Today news module on Yahoo!'s homepage over the course of 
several days in May 2009, totaling 45,811,883 separate events.

Each row in the dataset describes an interaction event (click or no click) of users shown a randomly chosen article. Each of these events contains the following information:

\begin{enumerate}
         \item The ID's of each of a varying subset of 19 to 25 articles selected by human editors from a pool of 217 articles.
         \item The ID of an article randomly chosen from the subset defined in 1. and positioned at the story position (that is, at the top of the Yahoo!'s website's "Today" 
	 segment).
         \item Six features per article for each of the articles shown.
         \item Six user features (with a distinct user for each event).
         \item Whether or not a user clicked on the article at the story position.
\end{enumerate}

That is, for each event $t$ an article represents one of $A$ actions (that is, one of the 271 articles observed within the course of the 10 days covered by the dataset) with 
$\mathbb{R}^6$ features $X_{t,a}$ per arm, and another $\mathbb{R}^6$ features $X_{t,u}$ per unique visitor. Together, the flattened outer product of the user and article 
feature vector creates a $\mathbb{R}^{36}$ feature vector $X_t$ for each user and article pair with outcome value or reward $r_t$ click (1) or no click (0). For the further 
details on the data structure and the general setup of the experiment, we refer the reader to \cite{Chu2009} and to the original \cite{Li2010} paper.

\subsection{Data import} \label{dataimp}

As the Yahoo data is too large to fit into memory, we imported most\footnote{The first two CSV files representing the first two days of the Yahoo! dataset are somewhat 
irregular, as they contain articles with more than six features. We, therefore, decided to leave these two CSV files out of our import, resulting in 37,450,196 imported 
events, instead of the 45,811,883 events used in the original paper.} of the dataset's CSV files into a \pkg{MonetDB} \citep{IdreosGNMMK12} instance---a fast, open source 
column-oriented database management system with excellent \proglang{R} support\footnote{MonetDB can be downloaded at https://www.monetdb.org/}. The import script, example 
import scripts for several other databases (\pkg{MySQL}, \pkg{SQLite}, \pkg{Postgresql}) and all other source code related to this replication can be found in the package's 
\code{demo/replication_li_2010} directory.

\subsection{Custom bandit and policies} \label{custom}

With the Yahoo! data imported into the MonetDB server, the next step was to create a custom offline \code{YahooBandit} plus seven \code{Policy} subclasses implementing the 
policies described in the \cite{Li2010} paper. Though most of these policies were already implemented in \pkg{contextual}, the fact that only a subset of all 271 articles or 
arms are shown to a visitor at a time meant we needed to make some minor changes to \pkg{contextual}'s exisiting classes to make the policies run smoothly on a continually 
shifting pool of active arms.

To facilitate these shifting arms, \code{YahooBandit} makes use of an \code{self$arm_lookup} table listing all 271 arms. This table enables the bandit to look up the 
currently active arms' indexes from the shifting set of article ID's as specified in the dataset for each time step $t$, and return these indexes to the policies under 
evaluation:

\begin{Code}
    get_context = function(index) {
      ...
      # Retrieve the index of all arms this row/event.
      arm_indices_this_event  <- seq(10, 184, by = 7)
      article_ids             <- row[arm_indices_this_event]
      article_ids             <- article_ids[!is.na(article_ids)]
      article_ids             <- match(article_ids,self$arm_lookup)
      ...
      context <- list(
        k = self$k,
        d = self$d,
        unique = self$unique, # Indexes of disjoint arms (user features)
        shared = self$shared, # Indexes of shared arms (article features)
        arms = article_ids,   # Indexes of arms this event.
        X = X
      )
    }
\end{Code}

The policy classes then use this information to select and update only the currently active subset of arms. For instance, in \code{YahooEpsilonGreedyPolicy}'s 
\code{get_action()}:

\begin{Code}
    get_action = function(t, context) {
      if (runif(1) > self$epsilon) {
        # get the max of context$arms *currently in play*
        max_index          <- context$arms[max_in(theta$mean[context$arms])]
        self$action$choice <- max_index
      } else {
        # sample from the arms *currently in play*
        self$action$choice <- sample(context$arms, 1)
      }
      self$action
    }
\end{Code}

On completing the implementation of the aforementioned seven custom policy subclasses (\code{Random}, \code{EGreedy}, \code{EGreedySeg}, \code{LinUCBDis}, \code{LinUCBHyb}, 
\code{UCB1} and \code{UCB1Seg}\footnote{\code{EGreedyDis} and \code{EGreedyHyb} policies were too summarily described for us to be able to replicate them with confidence.}) 
we then assigned them to six simulations---one for each of the six (0, 30, 20, 10, 5 and 1 percent) levels of sparsity  defined in the original paper. This resulted in 
$7\times6=42$ Agents, which were then run on the offline dataset as follows:

\begin{Code}
simulations             <- 1
horizon                 <- 37.45e6
...
con <- DBI::dbConnect(MonetDB.R(), host=monetdb_host, dbname=monetdb_dbname,
                                   user=monetdb_user, password=monetdb_pass)

message(paste0("MonetDB: connection to '",dbListTables(con),"' succesful!"))

arm_lookup_table <-
  as.matrix(DBI::dbGetQuery(con, "SELECT DISTINCT article_id FROM yahoo"))

arm_lookup_table <- rev(as.vector(arm_lookup_table))

bandit <- YahooBandit$new(k = 217L, unique = c(1:6), shared = c(7:12),
                          arm_lookup = arm_lookup_table, host = monetdb_host,
                          dbname = monetdb_dbname, user = monetdb_user,
                          password = monetdb_pass, buffer_size = buffer_size)

agents <-
  list (Agent$new(YahooLinUCBDisjointPolicy$new(0.2),
                  bandit, name = "LinUCB Dis",  sparse = 0.99),
        Agent$new(YahooLinUCBHybridPolicy$new(0.2),
                  bandit, name = "LinUCB Hyb",  sparse = 0.99),
        Agent$new(YahooEpsilonGreedyPolicy$new(0.3),
                  bandit, name = "EGreedy",     sparse = 0.99),
        Agent$new(YahooEpsilonGreedySegPolicy$new(0.3),
                  bandit, name = "EGreedySeg",  sparse = 0.99),
        Agent$new(YahooUCB1AlphaPolicy$new(0.4),
                  bandit, name = "UCB1",        sparse = 0.99),
        Agent$new(YahooUCB1AlphaSegPolicy$new(0.4),
                  bandit, name = "UCB1Seg",     sparse = 0.99),
        ...
        Agent$new(YahooRandomPolicy$new(),
                  bandit, name = "Random"))

simulation <- Simulator$new(
    agents,
    simulations = simulations,
    horizon = horizon,
    do_parallel = TRUE,
    worker_max = worker_max,
    reindex = TRUE,
    progress_file = TRUE,
    include_packages = c("MonetDB.R"))

history  <- simulation$run()
...
\end{Code}

\subsection{Results} \label{rslts}

\begin{figure}[H]
  \centering
    \includegraphics[width=.99\textwidth]{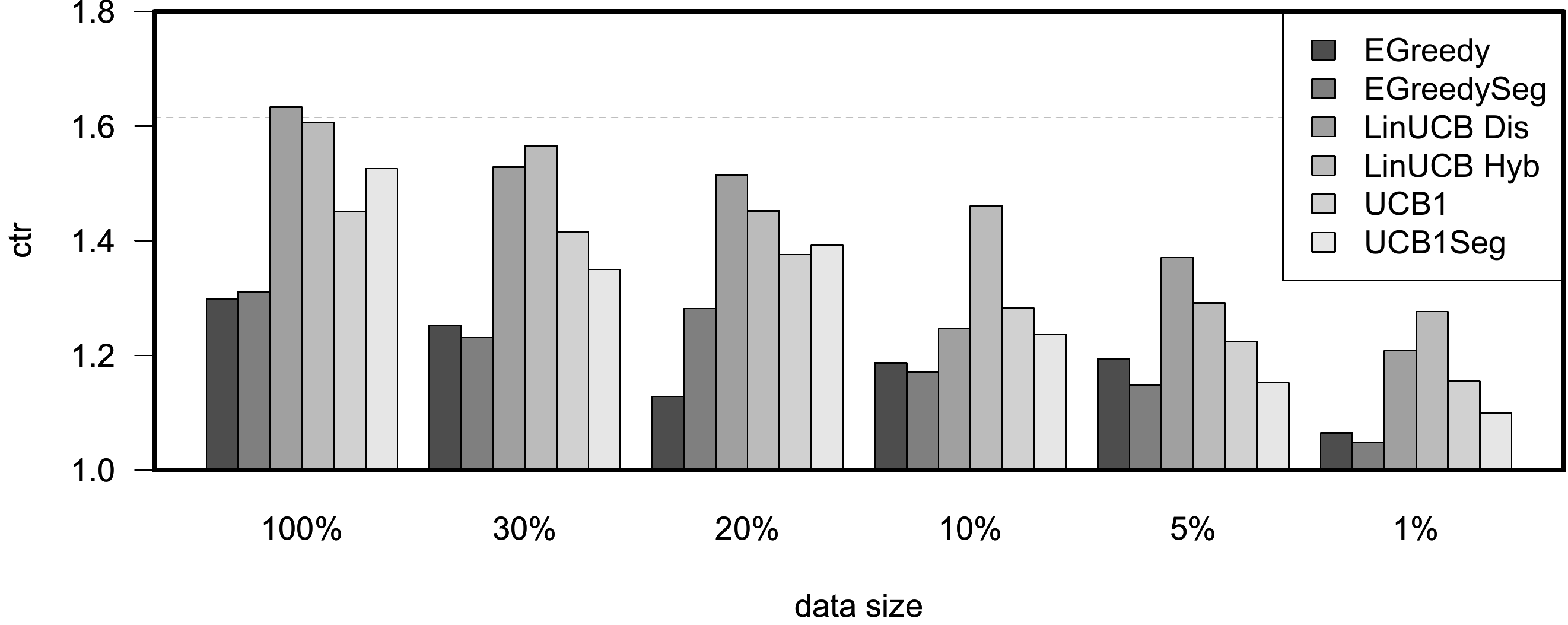}
      \caption{Replication of Figure 4b "CTRs in evaluation data with varying data sizes in the learning bucket." from \cite{Li2010}. The plots represent the click through 
      rates for different policies after one day of learning for each of six different levels of sparsity.}
      \label{fig:section_8_bar}
\end{figure}

We were able to complete the full $7\times6=42$ agent simulation over all of the 37,450,196 events in our database within 22 hours on a 64 core Intel Xeon Unbuntu server 
with 256GB of memory. We then proceeded to analyse the results of the first 4.7 million events (following the original paper, representing about a day worth of events) to 
reproduce \cite{Li2010}'s Figure 4b: "CTRs in evaluation data with varying data sizes in the learning bucket.". Just like the original paper, the replicated Figure 
\ref{fig:section_8_bar} reports each algorithm’s relative CTR for all of the defined data sparsity levels, that is, each algorithm’s CTR divided by the random policy’s CTR.

As can be observed in Figure \ref{fig:section_8_bar}, after one day of learning (on the Yahoo! dataset's day three), the conclusions of the original paper still stand. 
First, features again prove to be of use at all levels of sparsity, as LinUCB policies outperform the others consistently. Second, UCB policies generally outperform 
$\epsilon$-greedy ones. And third, Hybrid LinUCB again shows benefits when the data is small, as can be deduced from it doing better in the 1\% bucket. Still, as we started 
our simulation on the third day instead of the first, our results are close to, but not quite the same as those reported in the original paper. Particularly the third 
conclusion, that of the relative advantage of Hybrid LinUCB with sparse data, seems to be slightly less convincing in our Figure \ref{fig:section_8_bar}.

So we decided to run a simulation that continued to learn beyond the first day for the sparse (1\%) data condition to test whether Hybrid LinUCB's reported relative 
advantage would prove stable over time.

\begin{figure}[H]
  \centering
    \includegraphics[width=.99\textwidth]{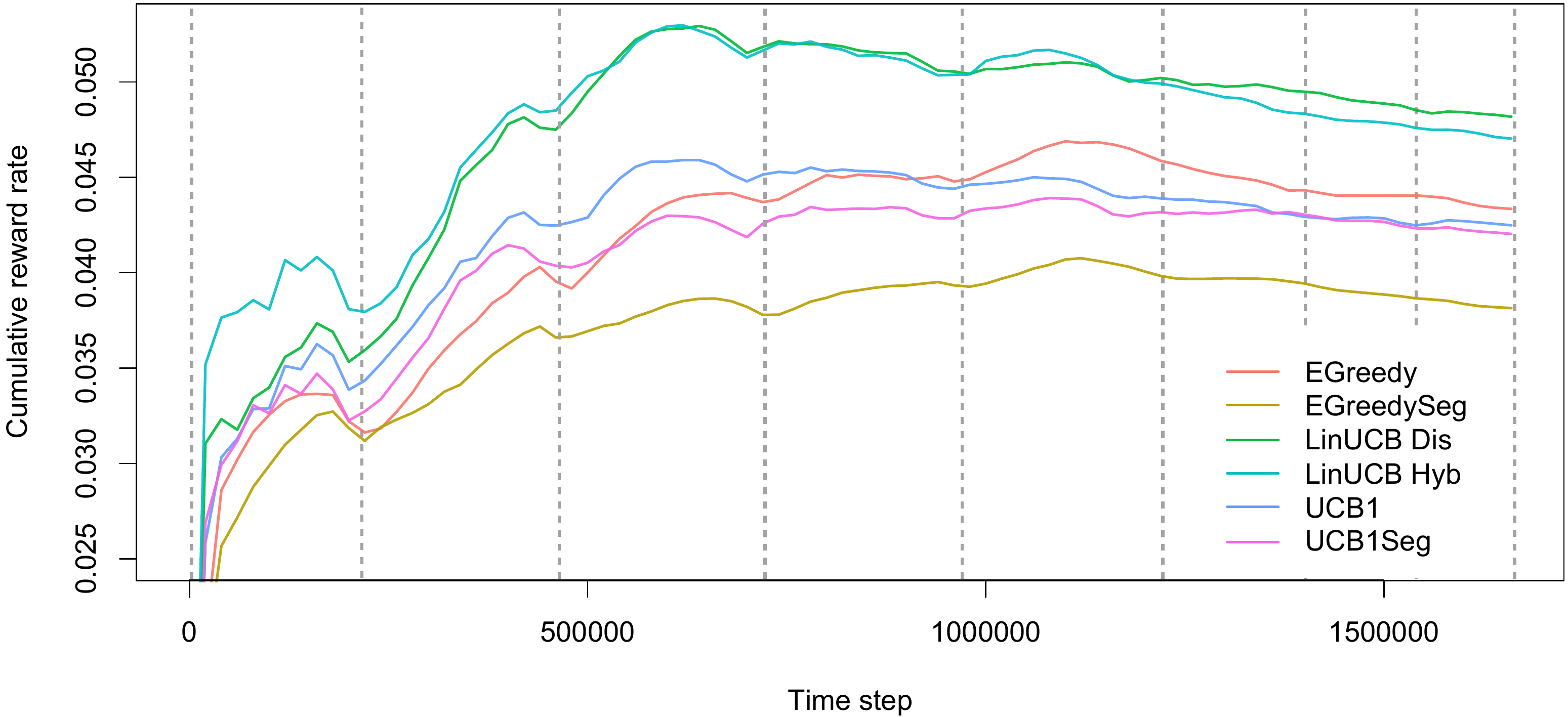}
      \caption{A plot of the cumulative reward rate (equals click-through rate) for EGreedy, EGreedySeg, LinUCB Dis, LinUCB Hyb, UCB1, and UCB1Seg policies over eight days 
      of events from the Yahoo dataset at 1\% sparsity. The dotted lines represent blocks consisting of 24 hours worth of data---the lines are not equidistant as the number 
      of data points of these 24 hours differs per block. The plot's sinusoidal wave pattern reflects visitors' overall average cyclical daily waxing and waning 
      click-through tendency.}
      \label{fig:section_8_plot}
\end{figure}

On closer examination of the resulting plot in Figure \ref{fig:section_8_plot}, it becomes clear that the policies did not settle after one day of training. For one, on 
inspection of the full span of about eight days of learning, Hybrid LinUCB's advantage over Disjoint LinUCB gradually diminishes and then reverses into an advantage for the 
Disjoint version. Also, surprisingly, over the full eight days, the $\epsilon$-greedy policy goes from the worst to third best policy overall, becoming the best context-free 
policy---clearly outperforming both context-free UCB policies. Though we intend to further analyze these discrepancies, for now, these results seem to pose questions for two 
out of three conclusions drawn in the original paper---leaving only the first outcome, the superiority of the contextual LinUCB in comparison to several context-free ones.  
Underlining the benefits of \pkg{contextual}, as it enabled us to replicate and confirm the original \cite{Li2010} paper, and then explore it further: extending 
\code{Bandit} and \code{Policy} classes, running simulations in parallel on an offline dataset and plotting and analyzing the results---within 48 hours.

\section{Discussion and future work} \label{future}

Statistical computational methods, in \proglang{R} or otherwise, are regularly made available through single-use scripts or basic, isolated code packages 
\citep{Gandrud2016}. Usually, such code examples are meant to give a basic idea of a statistical method, technique or algorithm in the context of a scientific paper 
\citep{Stodden2013}. Such code examples offer their scientific audience a first inroad towards the comparison and further implementation of their underlying methods 
\citep{Buckheit1995}. However, when a set of well-researched interrelated algorithms, such as MAB and CMAB policies, find growing academic, practical and commercial 
adoption, it becomes crucial to offer a more standardized and more accessible way to compare such methods and algorithms \citep{Mesirov2010}.

It is on that premise that we decided to develop the \pkg{contextual} \proglang{R} package---a package that would offer an open bandit framework with easily extensible 
bandit and policy libraries. To us, it made the most sense to create such a package in \proglang{R} \citep{RCore}, as \proglang{R} is currently the de facto language for the 
dissemination of new statistical methods, techniques, and algorithms \citep{Tippmann2015}---while it is at the same time finding ever-growing adoption in industry 
\citep{2012}. The resulting lively exchange of \proglang{R} related code, data, and knowledge between scientists and practitioners offers precisely the kind of 
cross-pollination that \pkg{contextual} intends to facilitate.

As the package is intended to be usable by practitioners, scientists and students alike, we started our paper with a general introduction to the (contextual) multi-armed 
bandit problem, followed by a compact formalization. We then demonstrated how our implementation flows naturally from this formalization, with \code{Agents} that cycle 
\code{Bandits} and \code{Policies} through four function calls: \code{get_context()}, \code{get_action()}, \code{get_reward()} and \code{set_reward()}. Next, we evaluated 
some of \pkg{contextual}'s built-in policies, delved deeper into \pkg{contextual}'s class structure, extended \pkg{contextual}'s \code{Bandit} and \code{Policy} 
superclasses, demonstrated how to evaluate Policies on offline datasets, and, finally replicated a frequently cited CMAB paper.

Though the package is fully functional and we expect no more changes to its core architecture and API, there is ample room to further improve and extend \pkg{contextual}. We 
intend to further expand \pkg{contextual}'s documentation and tests. We expect to include more bandit paradigms, such as dueling and combinatorial bandits. We expect to add 
other offline bandit types, such as a doubly robust bandit \citep{Dudik2011}. We are interested in growing our policy library---possibly by creating a separate repository 
where both existing and new CMAB policies are shared, evaluated and compared. Finally, we hope that the package will find an active community of users and developers, 
thereby introducing more and more people to the refined sequential decision strategies offered by contextual bandit policies and algorithms.

\bibliography{jss}

\newpage

\section{Appendix A: UML diagrams} \label{uml}

\begin{figure}[H]
  \centering
    \includegraphics[width=.99\textwidth]{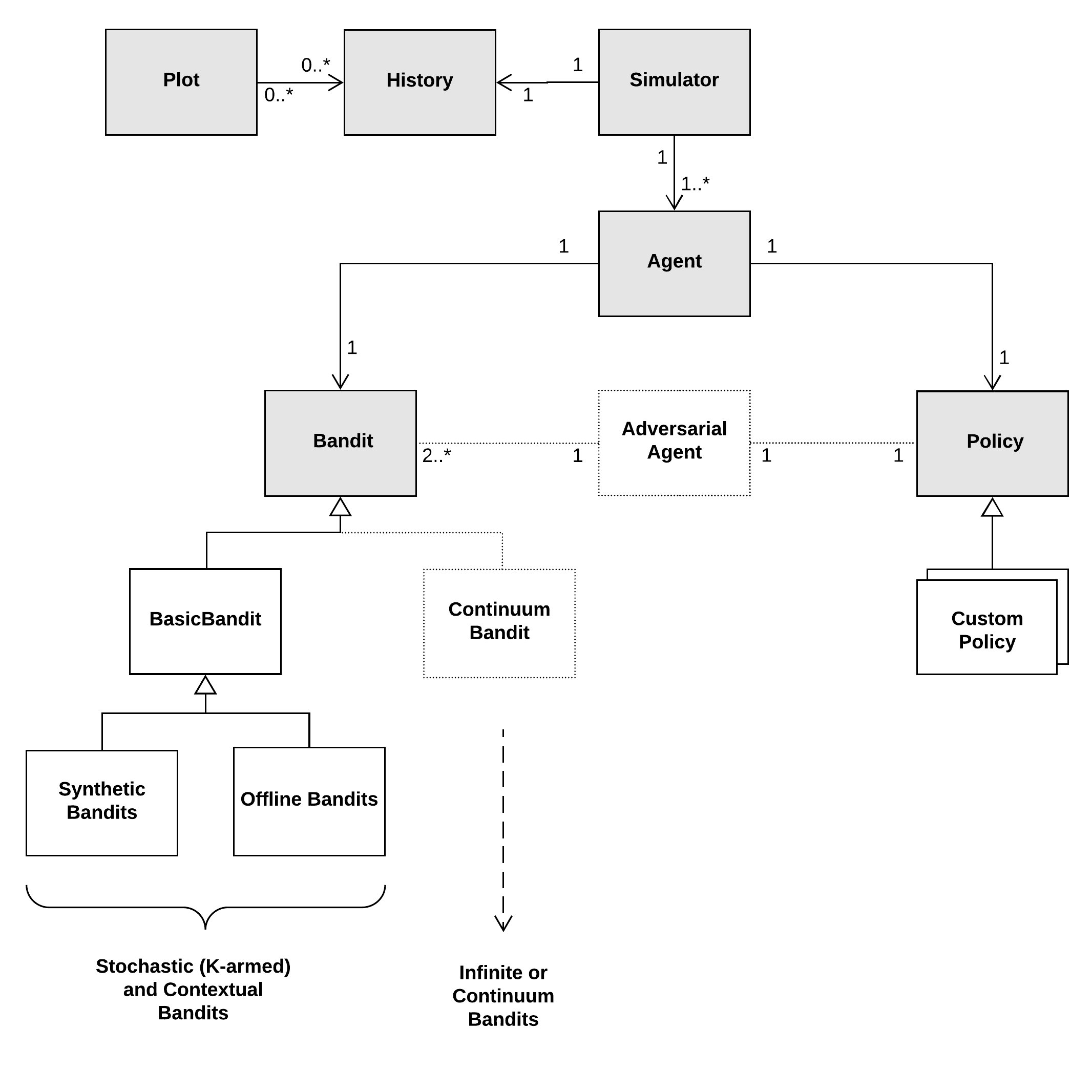}

      \caption{\pkg{contextual} UML Class Diagram}
          \label{fig:contextual_class}
\end{figure}

\begin{figure}[H]
  \centering
    \includegraphics[width=.99\textwidth]{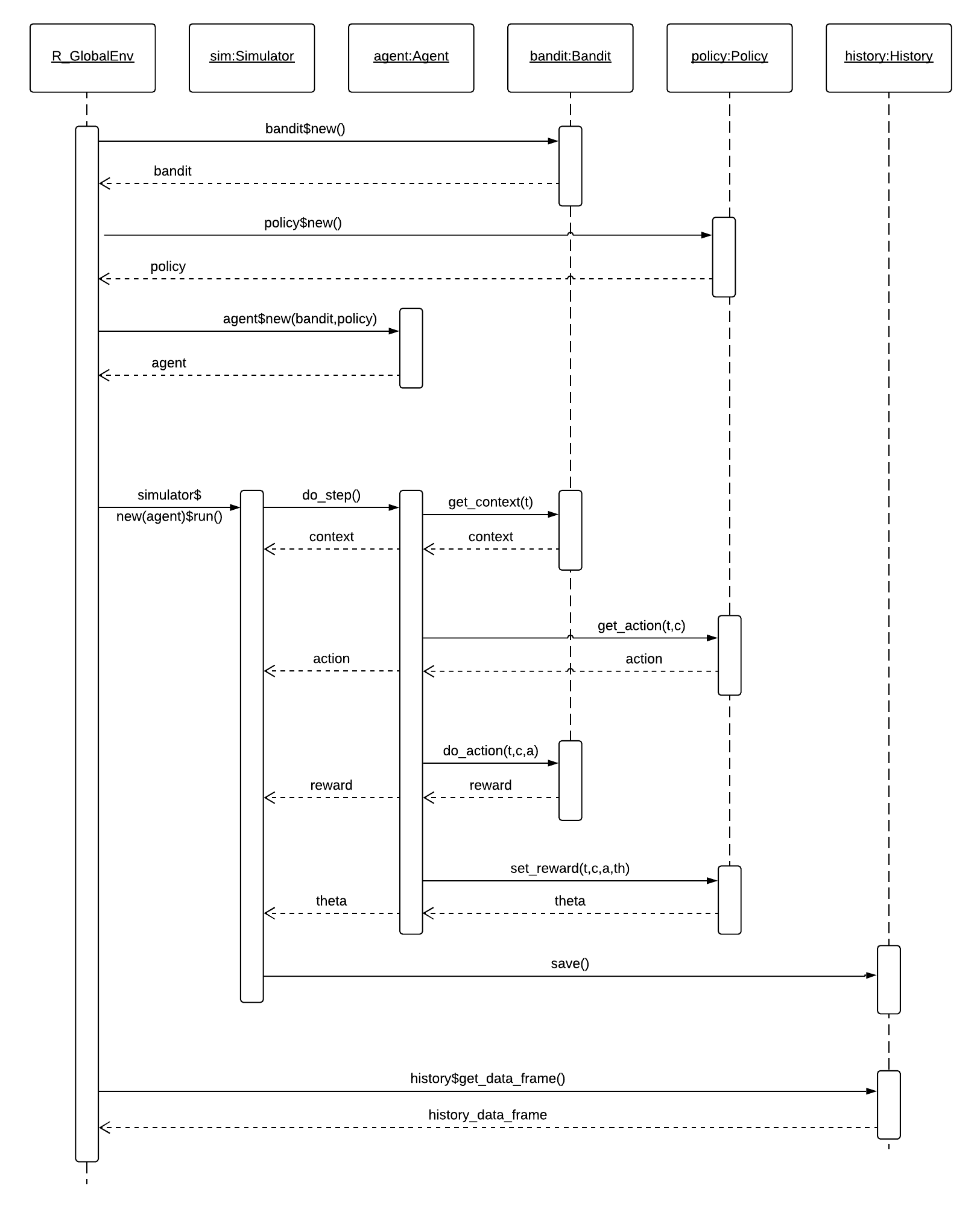}

      \caption{\pkg{contextual} UML Sequence Diagram}
      \label{fig:contextual_sequence}
\end{figure}

\end{document}